\documentclass[journal]{IEEEtran}
\usepackage{hyperref}
\usepackage{multirow}
\usepackage{booktabs}

\ifCLASSINFOpdf
  \usepackage[pdftex]{graphicx}
\else
\fi
\usepackage{array}

\ifCLASSOPTIONcompsoc
 \usepackage[caption=false,font=normalsize,labelfont=sf,textfont=sf]{subfig}
\else
 \usepackage[caption=false,font=footnotesize]{subfig}
\fi
\begin{document}
%
\title{DUNet: A deformable network for retinal vessel segmentation}
%
%
%

\author{Qiangguo~Jin,
\and
Zhaopeng~Meng,
\and 
Tuan D. Pham,
\and
Qi~Chen,
\and
Leyi~Wei,
\and
and~Ran~Su

\thanks{Manuscript submitted Oct. 18, 2018. (Corresponding author: Ran Su).}
\thanks{Qiangguo Jin is with School of Computer Software, College of Intelligence and Computing, Tianjin University, Tianjin, China (e-mail: qgking@tju.edu.cn).}
\thanks{Zhaopeng Meng is with School of Computer Software, College of Intelligence and Computing, Tianjin University, Tianjin, China (e-mail: mengzp@tju.edu.cn).}
\thanks{Tuan D. Pham is with Department of Biomedical Engineering, Linkoping University, Sweden (e-mail: tuan.pham@liu.se). }
\thanks{Qi Chen is with School of Computer Software, College of Intelligence and Computing, Tianjin University, Tianjin, China (e-mail: joannaxiaoqi@tju.edu.cn). }
\thanks{Leyi Wei is with School of Computer Science and Technology, College of Intelligence and Computing, Tianjin University, Tianjin, China (e-mail: weileyi@tju.edu.cn)}
\thanks{Ran Su is with School of Computer Software, College of Intelligence and Computing, Tianjin University, Tianjin, China (e-mail: ran.su@tju.edu.cn).}
}

\markboth{Journal of \LaTeX\ Class Files,~Vol.~14, No.~8, August~2015}%
{Shell \MakeLowercase{\textit{et al.}}: Bare Demo of IEEEtran.cls for IEEE Journals}
%



\maketitle

\begin{abstract}
Automatic segmentation of retinal vessels in fundus images plays an important role in the diagnosis of some diseases such as diabetes and hypertension. In this paper, we propose Deformable U-Net (DUNet), which exploits the retinal vessels' local features with a U-shape architecture, in an end to end manner for retinal vessel segmentation. Inspired by the recently introduced deformable convolutional networks, we integrate the deformable convolution into the proposed network. The DUNet, with upsampling operators to increase the output resolution, is designed to extract context information and enable precise localization by combining low-level feature maps with high-level ones. Furthermore, DUNet captures the retinal vessels at various shapes and scales by adaptively adjusting the receptive fields according to vessels' scales and shapes. Three public datasets DRIVE, STARE and CHASE\_DB1 are used to train and test our model. Detailed comparisons between the proposed network and the deformable neural network, U-Net are provided in our study. Results show that more detailed vessels are extracted by DUNet and it exhibits state-of-the-art performance for retinal vessel segmentation with a global accuracy of 0.9697/0.9722/0.9724 and AUC of 0.9856/0.9868/0.9863 on DRIVE, STARE and CHASE\_DB1 respectively. Moreover, to show the generalization ability of the DUNet, we used another two retinal vessel data sets, one is named WIDE and the other is a synthetic data set with diverse styles, named SYNTHE, to qualitatively and quantitatively analyzed and compared with other methods. Results indicates that DUNet outperforms other state-of-the-arts.
\end{abstract}


%
\IEEEpeerreviewmaketitle

\section{Introduction}
\label{sec:intro}
%
%
%
%
\IEEEPARstart{T}{he} morphological and topographical changes of retinal vessels may indicate some pathological diseases, such as diabetes and hypertension. Diabetic Retinopathy (DR) caused by elevated blood sugar levels, is a complication of diabetes in which retinal blood vessels leak into the retina, accompanying with the swelling of the retinal vessels~\cite{Smart2015A}. It must be noticeable if a diabetic patient appears in a swelling of the retinal vessels. Hypertensive Retinopathy (HR) is another commonly seen retina disease caused by high blood pressure~\cite{irshad_classification_2015}. An increased vascular tortuosity or narrowing of vessels can be observed in a patient with high blood pressure~\cite{cheung_retinal_2011}. Therefore, retinal blood vessels extracted from fundus images can be applied to the early diagnosis of some severe disease. This inspires the proposal of more accurate retinal blood vessel detection algorithms in order to facilitate the early diagnosis of pathological diseases.

However, the retinal blood vessels present extremely complicated structures, together with high tortuosity and various shapes~\cite{han_blood_2015}, which makes the blood vessel segmentation task quite challenging. Different approaches have been proposed for blood vessel detection. They are mainly divided into two categories: manual segmentation and algorithmic segmentation. The manual way is time-consuming and in high-demand of skilled technical staff. Therefore, automated segmentation of retinal vessels, which can release the intense burden of manual segmentation, is highly demanded. However, due to the uneven intensity distribution of the retinal vascular images, the subtle contrast between the target vessels and the background of the images, high complexity of the vessel structures, coupled with image noise pollution, it is quite challenging to segment the retinal blood vessels in an accurate and efficient way.

Deep learning has shown its excellence in medical imaging tasks. Recently, the Fully Conventional Neural Network (FCN) based network such as U-Net~\cite{ronneberger_u-net:_2015} has attracted more attention compared with the traditional Convolutional Neural Network (CNN) due to its ability to obtain a coarse-to-fine representation. In this study, we proposed an FCN-based network named Deformable U-Net (DUNet) that greatly enhances deep neural networks' capability of segmenting vessels in an end-to-end and pixel-to-pixel manner. It is designed to have a U-shape similar to U-Net~\cite{ronneberger_u-net:_2015} where upsampling operators with a large number of feature channels are stacked symmetrically to the conventional CNN, so context information is captured and propagated to higher resolution layers and thus a more precise segmentation is obtained. Furthermore, inspired by the recently proposed deformable convolutional networks (Deformable-ConvNet)~\cite{dai_deformable_2017}, we stacked deformable convolution blocks both in the encoder and decoder to capture the geometric transformations. Therefore, the receptive fields are adaptively adjusted according to the objects' scales and shapes and complicated vessel structures can be well detected. Deformable-ConvNet and U-Net are used for comparison. All the networks were trained from scratch and detailed analysis of the experimental results was provided. In the next section, we give a brief literature review of related work. Section~\ref{sec:method} explains the architecture of DUNet and systematic retinal blood vessel segmentation method. Deformable-ConvNet and U-Net are also introduced briefly in this section. Experimental results are presented in Section~\ref{sec:experimental}, where we evaluate the proposed method on three different retinal blood vessel datasets. Conclusions and discussions are given in Section~\ref{sec:conclusion}.

\begin{figure*}
\centering
\includegraphics[scale=0.6]{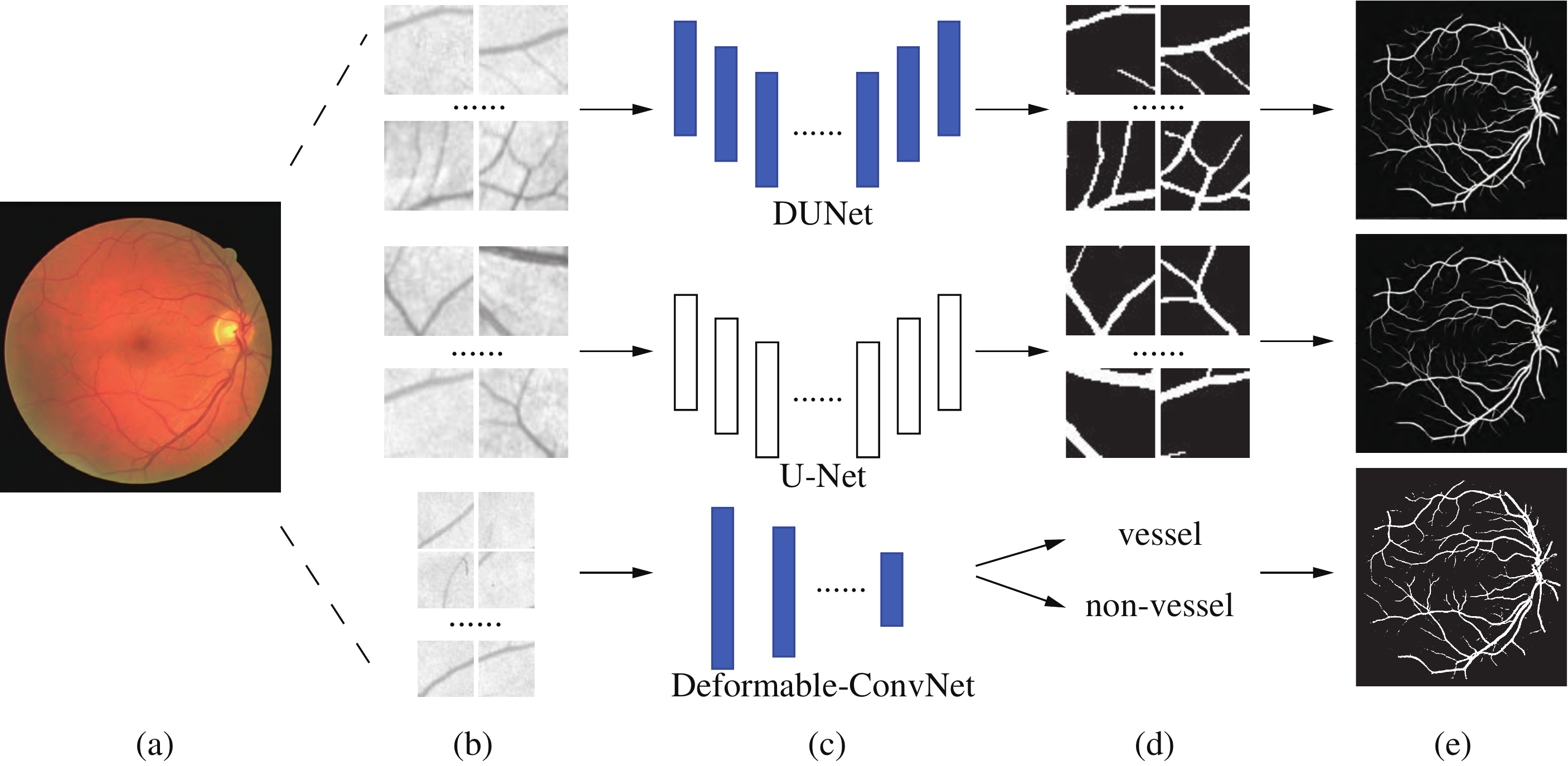}
\caption{The pipeline of the three networks. (a) Original image; (b) Training samples; (c) Snapshots of proposed DUNet and compared models. Note that blue blocks refer to deformable convolution and the white ones represent regular convolution; (d) Inference results; (e) Re-composition of segmentation results.}
\label{fig:syspipline}
\end{figure*}
\section{Related work}
\label{sec:related}
The goal of retinal blood vessel segmentation is to locate and identify retinal vessel structures in the fundus images. With the development of imaging technology, various intelligent algorithms have been applied to retinal vessel segmentation. According to the learning patterns, segmentation methods can be divided into supervised method and unsupervised method. Supervised learning learns from training data to generate a model and predict test data from that model, it automatically finds the probable category of data. Next, a brief overview of vessel segmentation from these two aspects is given.
\subsection{Unsupervised method}
The unsupervised method has no training samples in advance, and it constructs models directly in most cases. Zana et al. presented an algorithm based on mathematical morphology and curvature evaluation for the detection of vessel-like patterns in a noisy environment and they obtained an accuracy of 0.9377~\cite{zana_segmentation_2001}. Fraz et al. combined vessel centerlines detection and morphological bit plane slicing to extract vessel from retinal images~\cite{fraz_approach_2012}. Martinez-Perez et al. proposed a method to automatically segment retinal blood vessels based on multiscale feature extraction~\cite{martinez-perez_segmentation_2007}. Niemeijer et al. compared a number of vessel segmentation algorithms~\cite{niemeijer_comparative_2004}. According to this study, the highest accuracy of those compared algorithms reached 0.9416. Zhang et al. presented a retinal vessel segmentation algorithm using an unsupervised texton dictionary, where vessel textons were derived from responses of a multi-scale Gabor filter bank~\cite{zhang_retinal_2015}. A better performance would be obtained if a proper pre-processing was carried out. Hassan et al. proposed a method which combined mathematical morphology and k-means clustering to segment blood vessels~\cite{hassan_retinal_2015}. However, this method was not good at dealing with vessels of various widths. Tiny structures might be lost using this method. Oliveira et al. used a combined matched filter, Frangi's filter, and Gabor Wavelet filter to enhance the vessels~\cite{oliveira_unsupervised_2016}. They took the average of a few performance metrics to enhance the contrast between vessels and background. Jouandeau et al. presented an algorithm which was based on an adaptive random sampling algorithm~\cite{jouandeau_retinal_2014}. Garg et al. proposed a segmentation approach which modeled the vessels as trenches~\cite{garg_unsupervised_2007}. They corrected the illumination, detected trenches by high curvature, and oriented the trenches in a particular direction first. Then they used a modified region growing method to extract the complete vessel structure. A threshold of mean illumination level that was set empirically might bring bias in this method. Zardadi et al. presented a faster unsupervised method for automatic detection of blood vessels in fundus images~\cite{zardadi_unsupervised_2016}. They enhanced the blood vessels in various directions; Then they presented an activation function on cellular responses; Next, they classified each pixel via an adaptive thresholding algorithm; Finally, a morphological post-processing was carried out. However, several spots were falsely segmented into vessels which affected the final performance of the algorithm.

\subsection{Supervised method}
Different from unsupervised learning, supervised learning requires hand-labeled data in order to build an optimally predictive model. All the inputs are mapped to the corresponding outputs using the built model. It has been widely applied to the segmentation tasks. In order to reach the goal of segmentation, two processors are needed: One is an extractor to extract the feature vectors of pixels; The other one is a classifier to map extracted vectors to the corresponding labels. A number of feature extractors have been proposed, for instance, the Gabor filter~\cite{hamamoto_gabor_1998}, the Gaussian filter~\cite{nguyen_application_2011} etc. Various classifiers such as k-NN classifier~\cite{staal_ridge-based_2004}, support vector machine (SVM)~\cite{ricci_retinal_2007}~\cite{you_segmentation_2011}, artificial neural networks (ANN)~\cite{sinthanayothin_automated_1999}, AdaBoost~\cite{li_adaboost_2008} etc, have been proposed to deal with different tasks.

Supervised methods were used widely in retinal vessel segmentation. Aslani et al. proposed a new segmentation method which characterized pixels with a vector of hybrid features calculated via a different extractor. They trained a Random Forest classifier with the hybrid feature vector to classify vessel/non-vessel pixels~\cite{aslani_new_2016}. In order to simplify the model and increase the efficiency, the number of Gabor features should be reduced as small as possible. Marín et al. used Neural Network (NN) scheme for pixel classification and they computed a 7-D vector composed of gray-level and moment invariants-based features for pixel representation~\cite{marin_new_2011}. Yet the calculation cost was high and needed to be optimized.

For these traditional supervised methods, what features are used for classification greatly influence the final results of the prediction. However, they are often defined empirically, which requires the human intervention and may cause bias. Therefore, an automated and effective feature extractor is highly demanded to achieve higher efficiency.

Deep learning is an architecture referring to an algorithm set which can solve the image, text and other tasks based on backpropagation and multi-layer neural network. One of the most significant contributions of deep learning is that it can replace handcrafted features with features automatically learned from deep hierarchical feature extraction method~\cite{song_hierarchical_2013}.

In a number of fields such as image processing, bioinformatics, and natural language processing, various deep learning architectures such as Deep Neural Networks, Convolutional Neural Networks, Deep Belief Networks and Recurrent Neural Networks have been widely used and have shown that they could produce state-of-the-art results on various tasks. Recently, there are some studies that investigated the vessel segmentation problems based on deep learning. Wang et al. preprocessed the retinal vessel images and then combined two superior classifiers, Convolutional Neural Network (CNN) and Random Forest (RF) together to carry out the segmentation~\cite{wang_hierarchical_2015}. Fu et al. used the deep learning architecture, formulated the vessel segmentation to a holistically-nested edge detection (HED) problem, and utilized the fully convolution neural networks to generate vessel probability map~\cite{fu_retinal_2016}. Maji et al. used a ConvNet-ensemble based framework to process color fundus images and detect blood vessels~\cite{maji_ensemble_2016}. Jiang et al. proposed a method which defined and computed pixels as primary features for segmentation, then a Neural Network (NN) classifier was trained using selected training data~\cite{jiang_supervised_2015}. In this method, each pixel was represented by an 8-D vector. Then the unlabeled pixels were classified based on the vector. Azemin et al. estimated the impact of aging based on the results of the supervised vessel segmentation using artificial neural network~\cite{azemin_supervised_2014}. It showed that different age groups affected different aspect of segmentation results. Liskowski et al. proposed a supervised segmentation architecture that used a Deep Neural Network with a large training dataset which was preprocessed via global contrast normalization, zero-phase whitening, geometric transformations and gamma corrections~\cite{liskowski_segmenting_2016}. And the network classified multiple pixels simultaneously using a variant structured prediction method. Fu et al. regarded the segmentation as a boundary detection problem and they combined the Convolution Neural Networks (CNN) and Conditional Random Field (CRF) layers into an integrated deep network to achieve their goal~\cite{fu_deepvessel:_2016}.

Overall, it is expected that deep learning approaches can overcome the difficulties existed in the traditional unsupervised and supervised methods. In our study, we developed a systematic framework using the fully convolutional based methods to finish the effective and automatic segmentation task of retinal blood vessels.

\section{Methodology}
\label{sec:method}
The goal of our work is to build deep learning models to segment retinal vessels in fundus images. Inspired by U-Net~\cite{ronneberger_u-net:_2015} and deformable convolutional network (Deformable-ConvNet)~\cite{dai_deformable_2017}, we propose a new network named Deformable U-Net (DUNet) for retinal vessel segmentation task. The proposed approach is designed to integrate the advantages of both deformable unit and U-Net architecture. We will introduce our proposed method in details while giving a brief explanation of the two other networks as well.

Fig.~\ref{fig:syspipline} shows an overview of the proposed DUNet, U-Net and Deformable-ConvNet. The raw images are preprocessed and cropped into small patches to establish training and validation dataset. During contrastive experiments, different models will be set with corresponding patch size. Since DUNet and U-Net are both end-to-end deep learning frameworks for segmentation, a $48 \times 48$ patch size is used to trade off between computing complexity and efficiency. Meanwhile, Deformable-ConvNet is a model for vessel classification, a $29 \times 29$ patch size is chosen for training. After the inference of an image from the test dataset, all outputs from different models are re-composed to form a complete segmentation map respectively.

\begin{figure}
\centering
\includegraphics[scale=0.35]{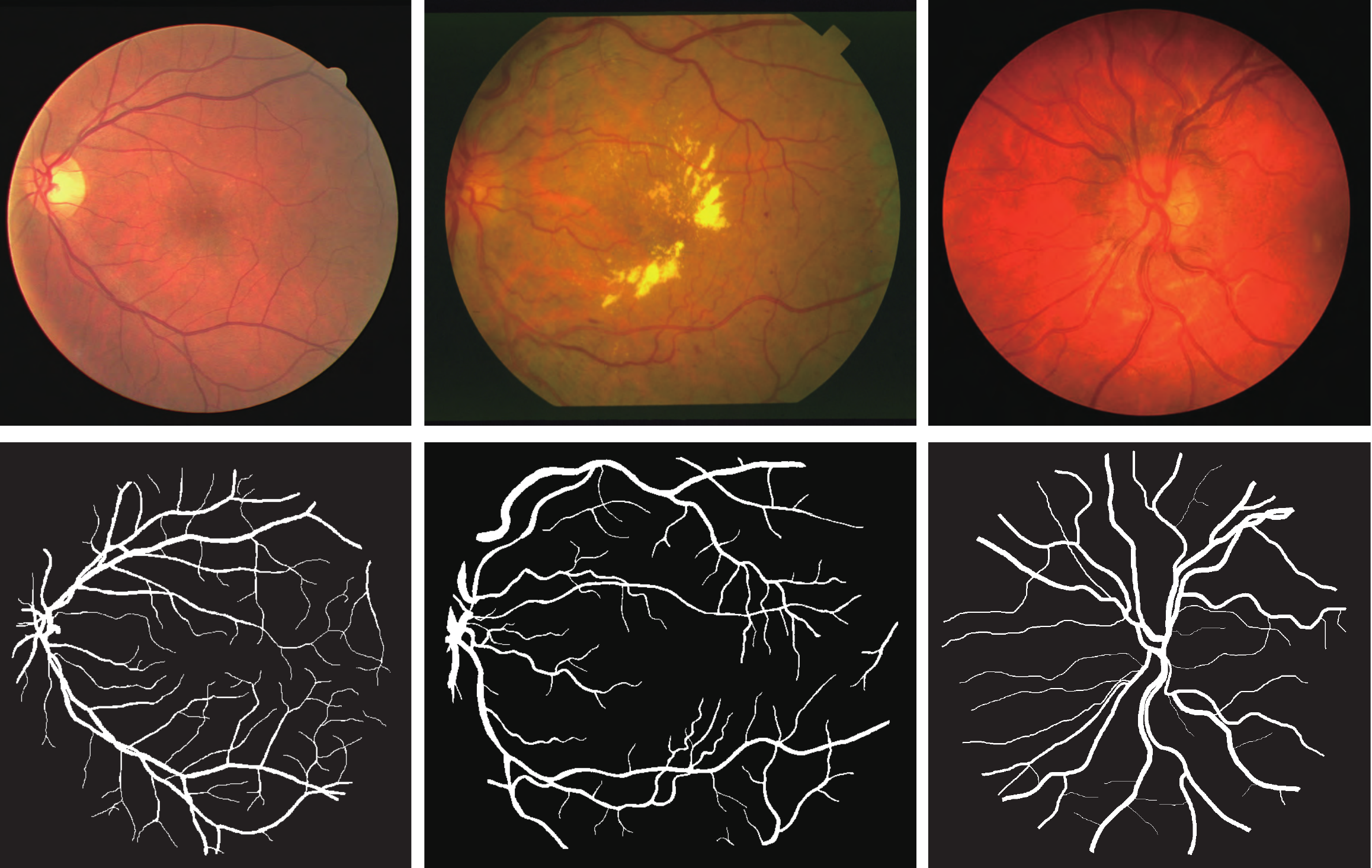}
\caption{Original retinal images (upper row) and corresponding ground truth (bottom row) examples from DRIVE, STARE and CHASE sequentially.}
\label{fig:raw_pics}
\end{figure}

\subsection{Datasets and material}
Performance was evaluated on three public datasets: DRIVE, STARE and CHASE\_DB1 (CHASE) dataset. The DRIVE (Digital Retinal Images for Vessel Extraction) contains 40 colored fundus photographs which were obtained from a diabetic retinopathy (DR) screening program in the Netherlands~\cite{Staal2004Ridge}. The plane resolution of DRIVE is $565 \times 584$. STARE (Structured Analysis of the Retina) dataset, which contains 20 images, is proposed to assist the ophthalmologist to diagnose eye disease~\cite{hoover_locating_1998}. The plane resolution of STARE is $700 \times 605$. The CHASE dataset contains 28 images corresponding to two per patient for 14 children in the program Child Hear And Health Study in England~\cite{Owen2009Measuring}. The plane resolution of CHASE is $999 \times 960$. Experts' manual annotations of the vascular are available as the ground truth (Fig.~\ref{fig:raw_pics}).

\subsection{Image preprocessing and dataset preparation}
\begin{figure}
\centering
\includegraphics[scale=0.6]{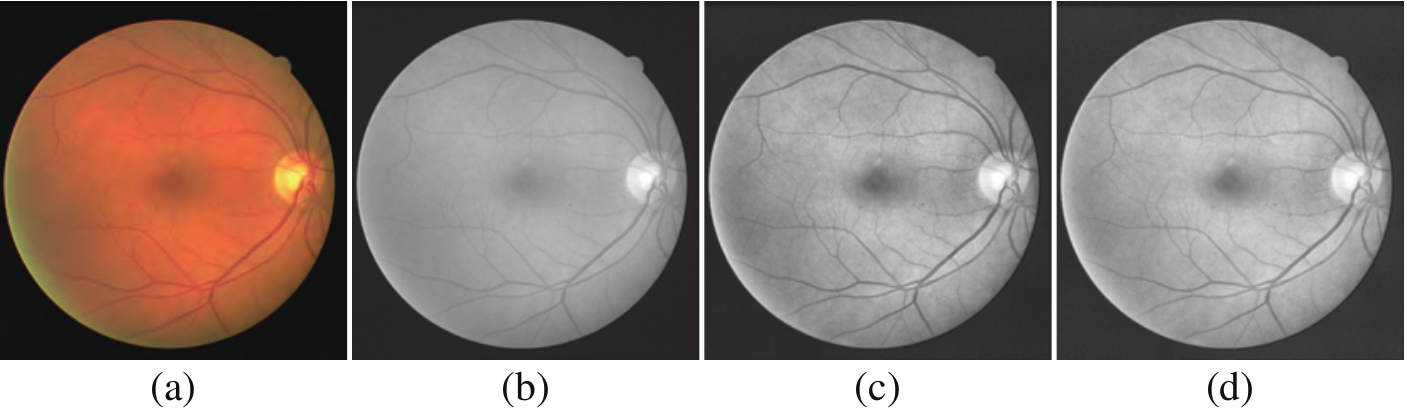}
\caption{Typical images after each preprocessing step. (a) Original image; (b) Normalized image; (c) Image after CLAHE operation; (d) Image after Gamma correction.}
\label{fig:preprocess}
\end{figure}

Deep neural network has the ability to learn from un-preprocessed image data effectively. While it tends to be much more efficient if appropriate preprocessing has been applied to the image data. In this study, three image preprocessing strategies were employed. Single channel images show the better vessel-background contrast than RGB images~\cite{soares_retinal_2006}. Thus, raw RGB images were converted into single channel ones. Normalization and Contrast Limited Adaptive Histogram Equalization~\cite{pizer_adaptive_1987} (CLAHE) were used over the whole data set to enhance the foreground-background contrast. Finally, gamma correction was introduced to improve the image quality much further. Intermediate images after each preprocessing step are shown in Fig.~\ref{fig:preprocess}.

\begin{figure}
\centering
\includegraphics[scale=0.23]{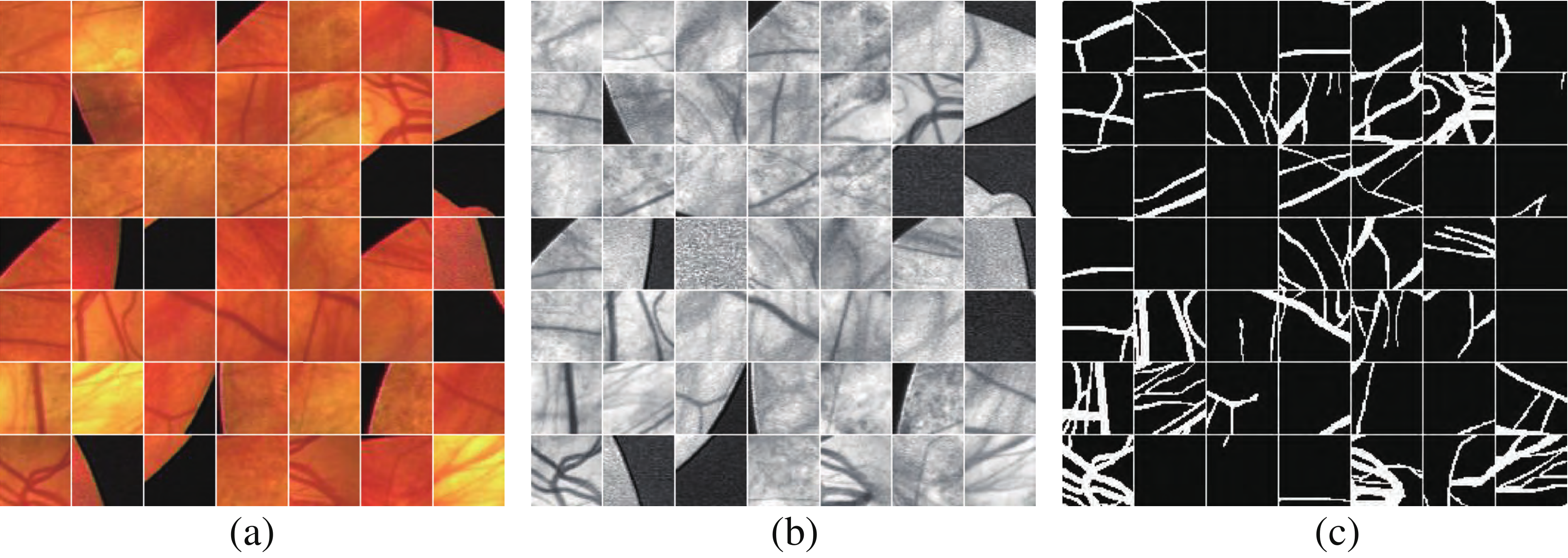}
\caption{Typical $48 \times 48$ patches selected for model training. (a) shows the patches from the original images; (b) shows the patches from the preprocessed image; (c) shows the patches from the corresponding ground truth.}
\label{fig:patches}
\end{figure}

To reduce overﬁtting problem, our models were trained on small patches which were randomly extracted from the images. In order to reduce the calculation complexity and ensure the surrounding local features, we set the size of the patch to $48 \times 48$ for DUNet and U-Net. The corresponding label for that patch was decided based on the ground truth images (Fig.~\ref{fig:patches}).

All the datasets were divided into training set, validation set, and test set. The training set is used for adjusting the weights. Validation set is used for selecting the best weight while test set is used for performance evaluation. For DRIVE dataset, 20 images were used for training and validating purpose and the rest for testing. Since no splitting of training or test is provided for STARE/CHASE, we manually separated the first 10/14 images for training and validating and the remaining 10/14 for testing. From each training/validating image on DRIVE, 10000 patches were randomly sampled including 8,000 for training and 2000 for validating. From each training/validating image on STARE/CHASE, 20000/15000 patches were randomly sampled including 16000/12000 for training and 4000/3000 for validating. Therefore, DRIVE and STARE both had 160000 patches as training set and 40000 patches as validation set. Meanwhile, CHASE had 168000 patches as training set and 42000 patches as validation set. The test set consists of the whole rest images. Since the capacity of patch dataset is large enough, data augmentation is not token into consideration.

\subsection{Deformable U-Net (DUNet)}
Inspired by U-Net~\cite{ronneberger_u-net:_2015} and deformable convolutional network (Deformable-ConvNet)~\cite{dai_deformable_2017}, we proposed a network, named Deformable U-Net (DUNet) for retinal vessel segmentation task. The proposed network has a U-shaped architecture with encoders and decoders on two sides, and the original convolutional layer was replaced by the deformable convolutional block. The new model is trained to integrate the low-level feature with the high-level features, and the receptive field and sampling locations are trained to adaptive to vessels' scale and shape, both of which enable precise segmentation. DUNet builds on top of U-Net and uses the deformable convolutional block as encoding and decoding unit.

\begin{figure*}
\centering
\includegraphics[scale=0.5]{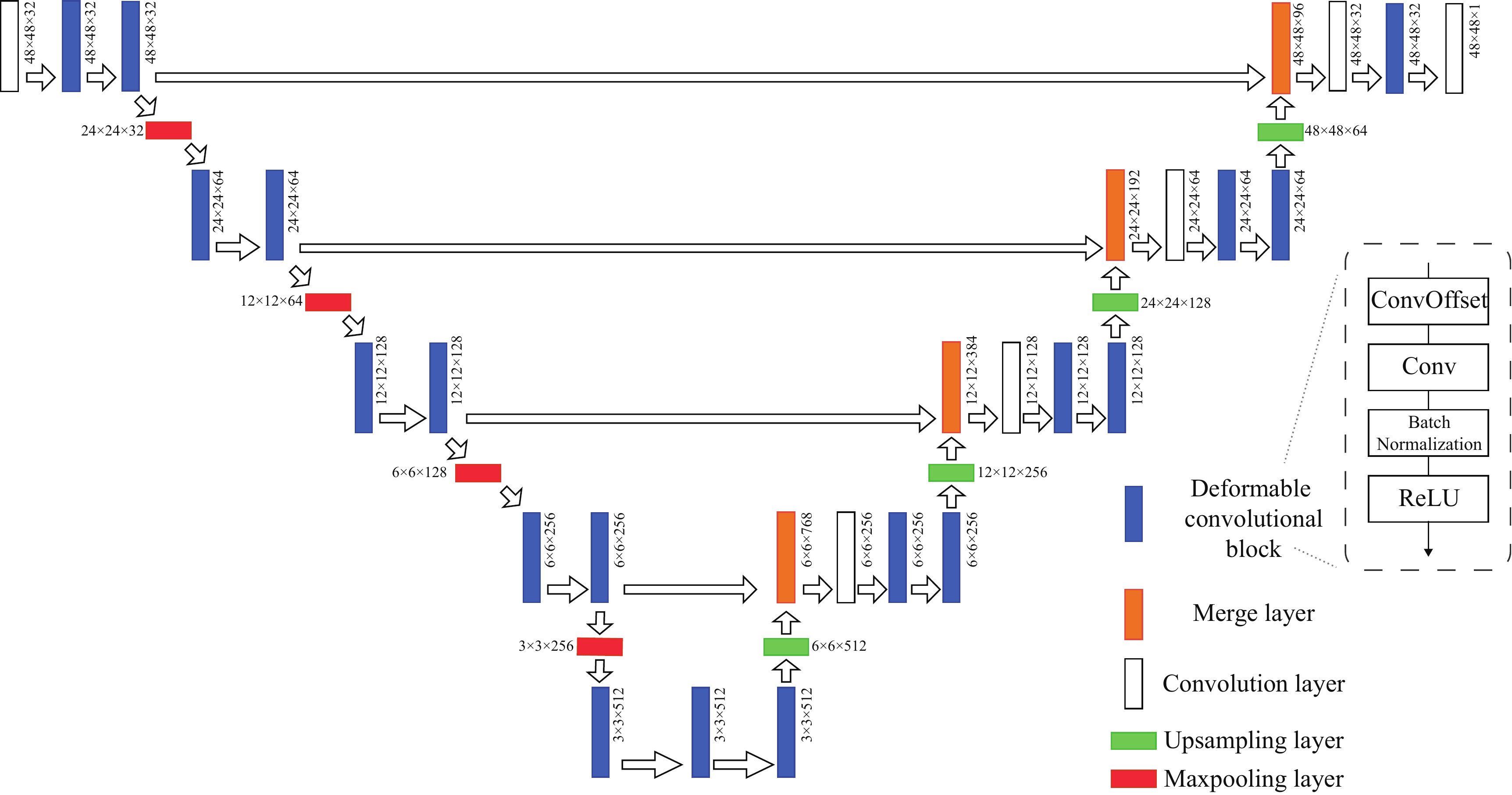}
\caption{DUNet architecture with convolutional encoder and decoder using deformable convolutional block based on U-Net architecture. Output size of feature map is listed beside each layer.}
\label{fig:DUNetpipline}
\end{figure*}

Fig.~\ref{fig:DUNetpipline} illustrates the network architecture. Detailed design of the deformable convolutional block is shown in the dashed window. The architecture consists of a convolutional encoder (left side) and a decoder (right side) in a U-Net framework. In each encoding and decoding phase, deformable convolutional blocks are used to model retinal vessels of various shapes and scales through learning local, dense and adaptive receptive fields. Each deformable convolutional block consists of a convolution offset layer, which is the kernel concept of deformable convolution, a convolution layer, a batch normalization layer~\cite{ioffe_batch_2015} and an activation layer. During the decoding phase, we additionally insert a normal convolution layer after merge operation to adjust filter numbers for convolution offset layer. With this architecture, DUNet can learn discriminative features and generate the detailed retinal vessel segmentation results.

\subsubsection{U-Net as the basic architecture}
Our U-Net architecture has an overall architecture similar to the standard U-Net, consisting of an encoder and a decoder symmetrically on the two sides of the architecture. The encoding phase is used to encode input images in a lower dimensionality with richer filters, while the decoding phase is designed to do the inverse process of encoding by upsampling and merging low dimensional feature maps, which enables the precise localization. Besides, in the upsampling part, a larger number feature channels are used in order to propagate the context to higher resolution layers. In order to solve the internal covariate shift problem and speed up the processing, a batch normalization layer was inserted after each unit.
\begin{figure}
\centering
\includegraphics[scale=0.45]{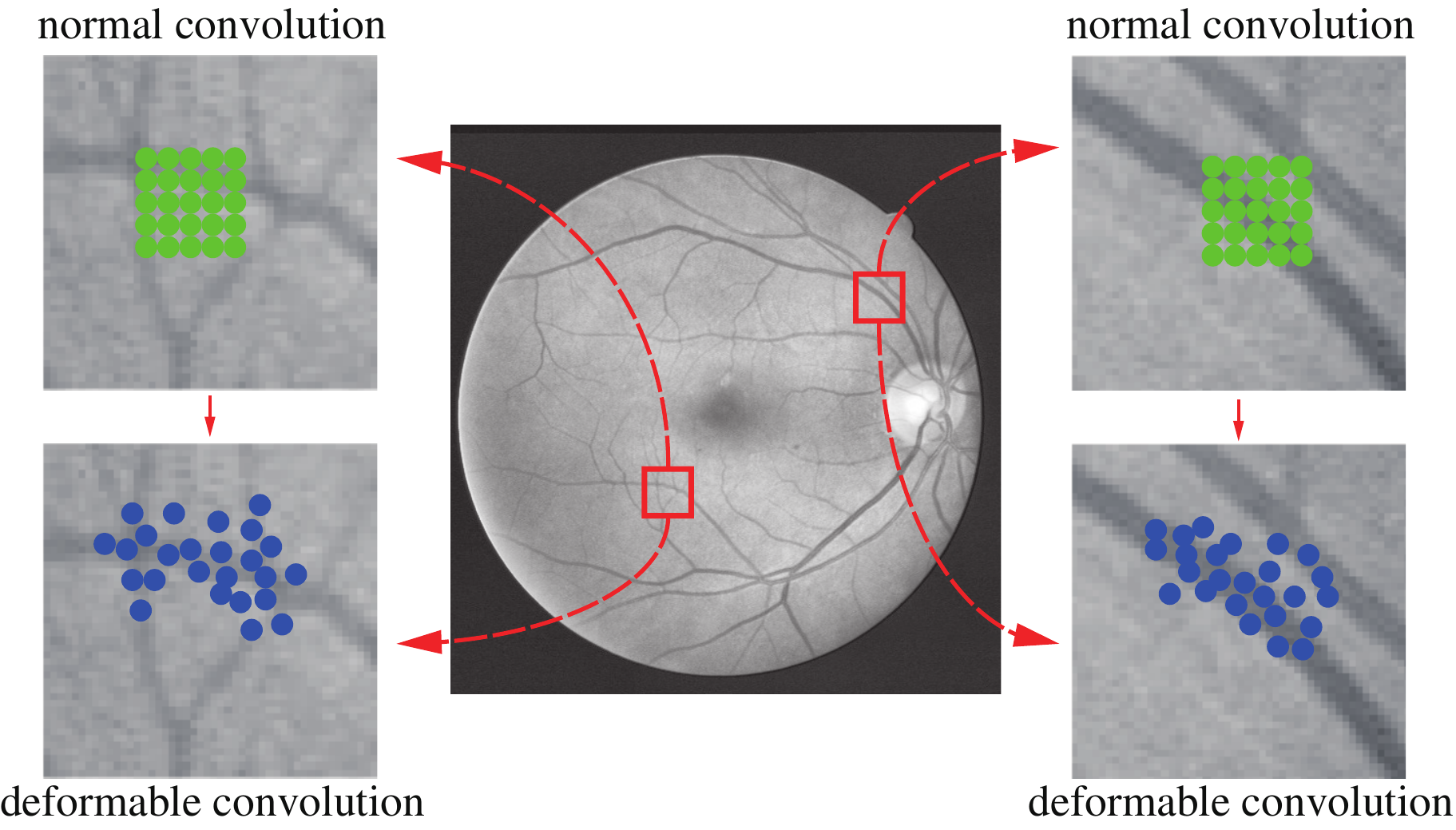}
\caption{Illustration of the sampling locations in $5 \times 5$ normal and deformable convolutions. The upper row stands for normal convolution and the corresponding deformable convolution is in the bottom row. Each sampling location has an offset to generate new sampling location.}
\label{fig:grid}
\end{figure}
\subsubsection{Deformable Convolutional Blocks}
A big challenge in vessel segmentation is to model the vessels with various shapes and scales~\cite{dai_deformable_2017}. Traditional methods such as the steerable filter~\cite{freeman_design_1991}, Frangi filter~\cite{frangi_muliscale_1998} exploit the vessel features through linear combination of responses at multiple scales or direction, which may bring bias. The deformable convolutional network (Deformable-ConvNet) solved this problem by introducing deformable convolutional layers and deformable ROI pooling layers into the traditional neural networks. We were inspired by the idea from Deformable-ConvNet that the various shapes and scales can be captured via deformable receptive fields, which are adaptive to the input features. Therefore, we integrated the deformation convolution into the proposed network.

In the deformable convolution, offsets were added to the grid sampling locations which are normally used in the standard convolution. The offsets were learned from the preceding feature maps produced by the additional convolutional layers. Therefore, the deformation is able to adapt to different scales, shapes, orientation, etc. We take the $5 \times 5$ deformable convolution as an example in Fig.~\ref{fig:grid}.

As Fig.~\ref{fig:grid} shows, for a $5 \times 5$ sized kernel with grid size 1, the normal convolution grid $G$ can be formalized as:
\setlength{\abovedisplayshortskip}{0pt}
\setlength{\belowdisplayshortskip}{0pt}
\begin{equation}
G=\left \{ (-2,2),(-2,-1),..., \right (2,1),(2,2) \}
\label{eq:grid}
\end{equation}

Thus, each location $m_0$ from output feature map $\mathbf{y}$ can be formalized as:
\setlength{\abovedisplayshortskip}{0pt}
\setlength{\belowdisplayshortskip}{0pt}
\begin{equation}
\mathbf{y}(m_{0})=\sum_{m_{i}\in G}\mathbf{w}(m_{i}) \cdot \mathbf{x}(m_{0}+m_{i})
\label{eq:y_m_0}
\end{equation}

Where $\mathbf{x}$ denotes the input feature map, $\mathbf{w}$ represents the weights of sampled value and $m_{i}$ means the locations in $G$. While in deformable convolution, normal grid $G$ is enhanced by the offset $\Delta m_{i}$, we have

\setlength{\abovedisplayshortskip}{0pt}
\setlength{\belowdisplayshortskip}{0pt}
\begin{equation}
\mathbf{y}(m_{0})=\sum_{m_{i}\in G}\mathbf{w}(m_{i}) \cdot \mathbf{x}(m_{0}+m_{i} +\Delta m_{i})
\label{eq:y_delta_m_0}
\end{equation}

\begin{figure}
\centering
\includegraphics[scale=0.45]{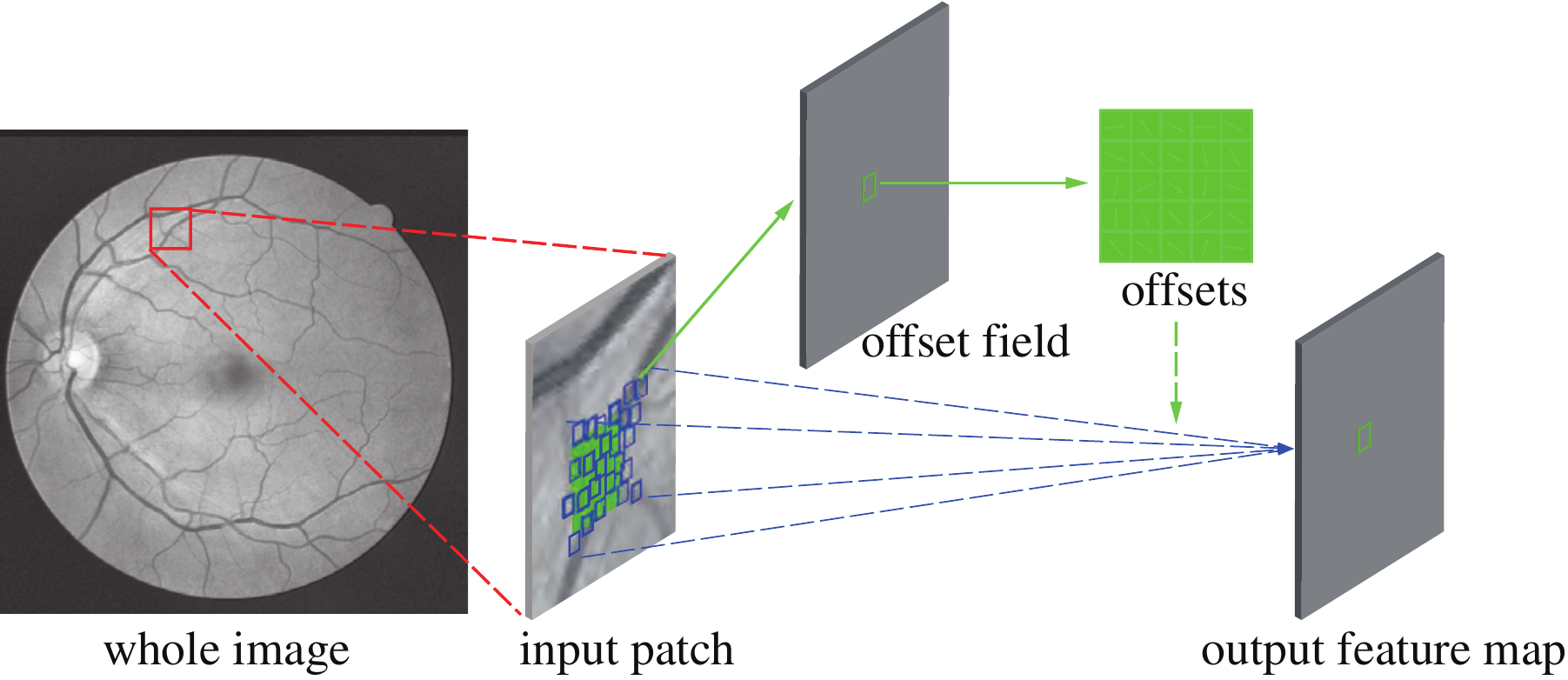}
\caption{Illustration of a $5 \times 5$ deformable convolution. Offset field comes from the input patches and features while the channel dimension is 2N corresponding to N 2D offsets. Deformable convolutional kernel has the same resolution as the current convolution layer. The convolution kernels and the offsets are learned at the same time.}
\label{fig:deformable}
\end{figure}

Because offset $\Delta m_{i}$ is usually not an integer, bilinear interpolation is applied to determine the value of the sampled points after migration. As mentioned above, the offset $\Delta m_{i}$ is learned by an additional convolution layer. This procedure is illustrated in Fig.~\ref{fig:deformable}. Compared to the regular U-Net, DUNet may incur some computation cost in order to perform in a more local and adaptive manner.

\subsection{Compare with U-Net and Deformable-ConvNet}
\begin{figure}
\centering
\includegraphics[scale=0.3]{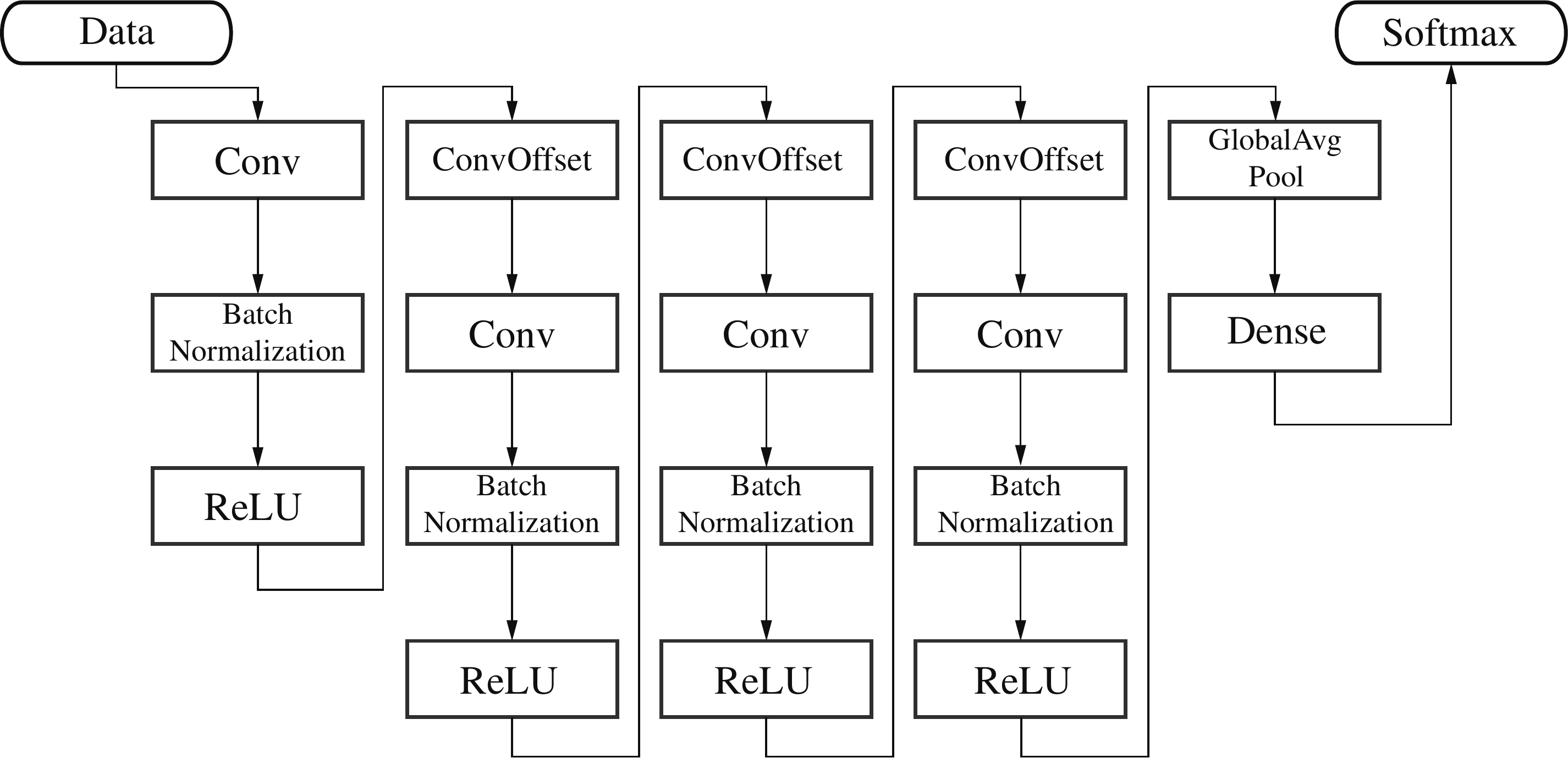}
\caption{The architecture of the Deformable-ConvNet. It is mainly composed of convolution layers (Conv), deformable convolutional layers (ConvOffset), batch normalization layers and activation layers (ReLU).
}
\label{fig:deformnet}
\end{figure}

We compared our proposed model with two state-of-the-art works. One is the normal U-Net, which we have introduced above; The other is the deformable convolutional network (Deformable-ConvNet). Deformable-ConvNet was originally introduced to distinguish whether a pixel belongs to vessel or not. In this model, vessel segmentation was considered as a classification task. A pixel's class can be determined based on its neighborhood defined as the patch centered on this pixel. For a selected pixel, which needs to be classified, we used pixel values in a patch centered on that selected pixel to capture the local information at a high level. In order to reduce the calculation complexity and to maximally capture the local features, the size of the patch was set to $29 \times 29$. The architecture of the Deformable-ConvNet is shown in Fig.~\ref{fig:deformnet}.

\subsection{Performance evaluation metrics}
We evaluated our model using several metrics: Accuracy (ACC), Positive Predictive Value (PPV), True Positive Rate (TPR), True Negative Rate (TNR) and the Area Under Curve (AUC) of Receiver Operating Characteristic (ROC). ACC is a metric for measuring the ratio between the correctly classified pixels and the total pixels in the dataset. PPV, which is also called precision, indicates the proportion of the true positive samples among all the predicted positive samples. TPR, also known as sensitivity, measures the proportion of positives that are correctly identified. TNR, or specificity, measures the proportion of negatives that are correctly identified. These metrics have the forms as following:

\setlength{\abovedisplayshortskip}{0pt}
\setlength{\belowdisplayshortskip}{0pt}
\begin{equation}
\mathrm{ACC}=\frac{\mathrm{TP}+\mathrm{TN}}{\mathrm{TP}+\mathrm{FP}+\mathrm{TN}+\mathrm{FN}}
\label{eq:accuracy}
\end{equation}

\setlength{\abovedisplayshortskip}{0pt}
\setlength{\belowdisplayshortskip}{0pt}
\begin{equation}
\mathrm{PPV}=\frac{\mathrm{TP}}{\mathrm{TP}+\mathrm{FP}}
\label{eq:PPV}
\end{equation}

\setlength{\abovedisplayshortskip}{0pt}
\setlength{\belowdisplayshortskip}{0pt}
\begin{equation}
\mathrm{TPR}=\frac{\mathrm{TP}}{\mathrm{TP}+\mathrm{FN}}
\label{eq:TPR}
\end{equation}

\setlength{\abovedisplayshortskip}{0pt}
\setlength{\belowdisplayshortskip}{0pt}
\begin{equation}
\mathrm{TNR}=\frac{\mathrm{TN}}{\mathrm{TN}+\mathrm{FP}}
\label{eq:SPC}
\end{equation}

Where TP represents the number of the true positive samples; TN stands for the number of the true negative samples; FP means the number of the false positive samples; FN means number of the false negative samples.

Additionally, performance was evaluated with F-measure ($\mathrm{F_{1}}$)~\cite{sasaki_truth_2007} and Jaccard similarity (JS)~\cite{jaccard_distribution_1901} to compare the similarity and diversity of testing datasets. Here GT refers to the ground truth and SR refers to the segmentation result.
\setlength{\abovedisplayshortskip}{0pt}
\setlength{\belowdisplayshortskip}{0pt}
\begin{equation}
\mathrm{F_{1}}=2\cdot \frac{\mathrm{PPV}\cdot \mathrm{TPR}}{\mathrm{PPV}+\mathrm{TPR}}
\label{eq:f_measure}
\end{equation}

\setlength{\abovedisplayshortskip}{0pt}
\setlength{\belowdisplayshortskip}{0pt}
\begin{equation}
\mathrm{JS}=\frac{\left | \mathrm{GT}\bigcap \mathrm{SR }\right |}{\left |\mathrm{GT}\bigcup \mathrm{SR}\right |}
\label{eq:jaccard}
\end{equation}

\section{Experimental result}
\label{sec:experimental}
The proposed DUNet has upsampling layers to increase output resolution. It enables propagation of the context information to the higher resolution layers and detection of vessels in various shapes and scales, thus presents an accurate segmentation result. In this section, we systematically compared the DUNet with Deformable-ConvNet and U-Net. We firstly show the results based on the validation set, which is used for parameter selection. Then the results on the test set are presented. We also briefly compared DUNet with some other recently published approaches, most of which are under deep neural network framework and the others are standard segmentation approaches. All experiments were conducted under the Tensorflow~\cite{abadi_tensorflow:_2016} and Keras~\cite{chollet_keras_2015} frameworks using an NVIDIA GeForce GTX 1080Ti GPU.

\subsection{Comparisons with Deformable-ConvNet and U-Net}
We compared the three models, Deformable-ConvNet, U-Net and DUNet based on the DRIVE, STARE and CHASE datasets. As described in Section~\ref{sec:method}, we split the data into training set, validation set, and test set. We trained the three models from scratch using the training set and initialized the weights with random values. We set the batch size to 60, total training epochs to 100, Adam as optimizer and binary cross-entropy as our loss function. To ensure a quick convergence and avoid overfitting, we used a dynamic method to set the learning rate values. The initial learning rate was set to 0.001. If the loss values remained stable after $m_{e}$ epochs, the learning rate was reduced 10 times. Additionally, the training process would be ceased if loss value stayed almost unchanged after $n_{e}$ epochs. Here $m_{e}$ and $n_{e}$ are set to 4 and 20 empirically. The validation accuracy and loss values were recorded during the training phase. Performance on validation dataset is reflected in Table~\ref{table:performance}.

\begin{table}[]
\caption{Performance of the three scratched-trained models on DRIVE, STARE and CHASE datasets}
\centering
\renewcommand\arraystretch{1.3}      
\renewcommand\tabcolsep{3.5pt}        
\begin{tabular}{ccccccc}
\toprule 
\multirow{2}{*}{Models} & \multicolumn{2}{c}{DRIVE}       & \multicolumn{2}{c}{STARE}         & \multicolumn{2}{c}{CHASE}   \\ \cline{2-7}
                        & ACC            & LOSS            & ACC             & LOSS            & ACC             & LOSS   \\ \hline
Deformable-ConvNet      & 0.9622         & 0.1101          & 0.9501          & 0.1593          & 0.9651     &0.0962\\ 
U-Net                   & 0.9648         & 0.1413          & \textbf{0.9573} & 0.2659          & 0.9664     &0.1366\\ 
\textbf{DUNet}                   & \textbf{0.9650} & \textbf{0.0919} & 0.9543         & \textbf{0.1477} & \textbf{0.9704}   &\textbf{0.0833}\\ 
\bottomrule 
\end{tabular}
\label{table:performance}
\end{table}

\begin{figure*}
\centering
\subfloat[]{\includegraphics[scale=0.3]{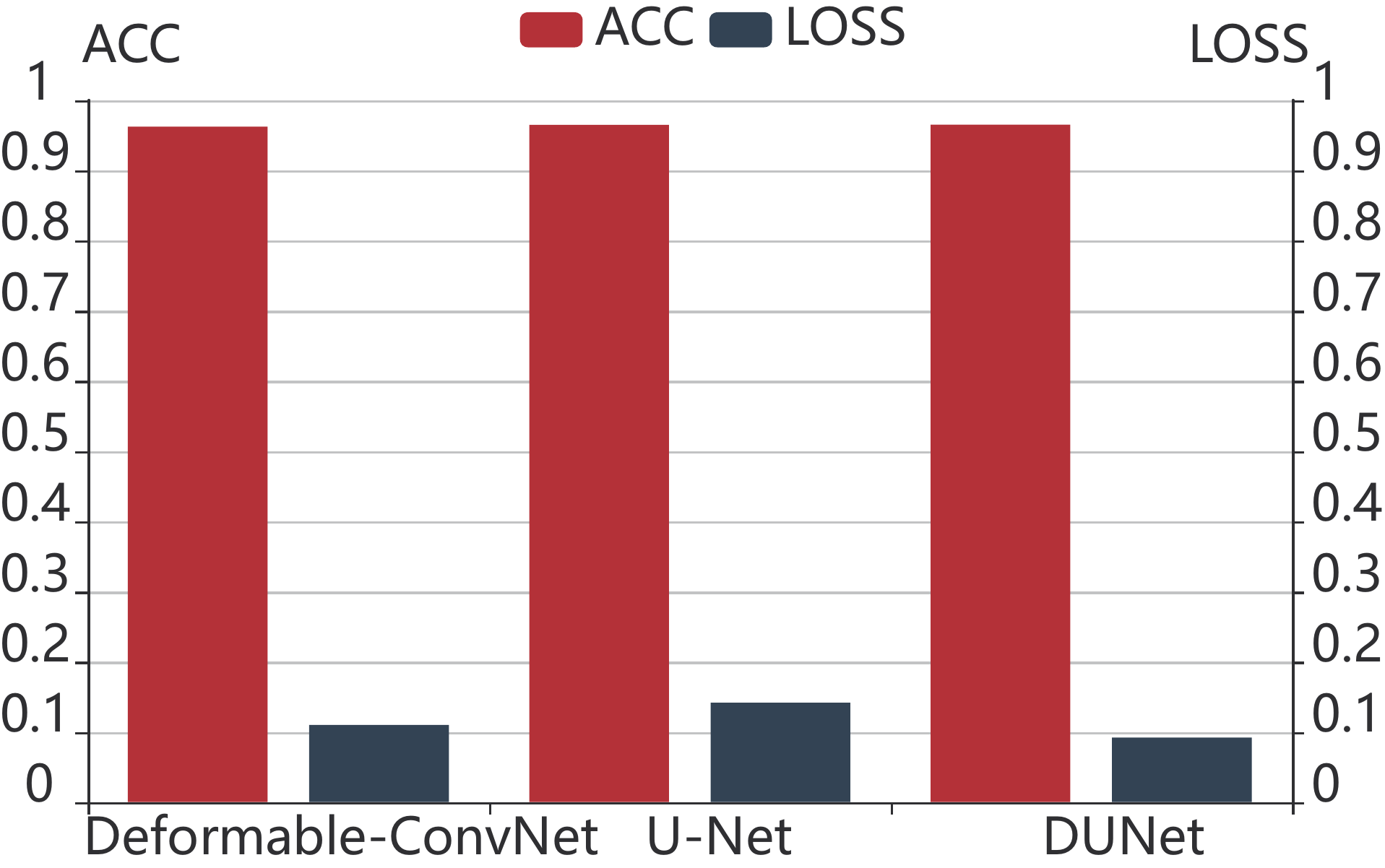}%
\label{drive_performance_svg}}
\hfil
\subfloat[]{\includegraphics[scale=0.3]{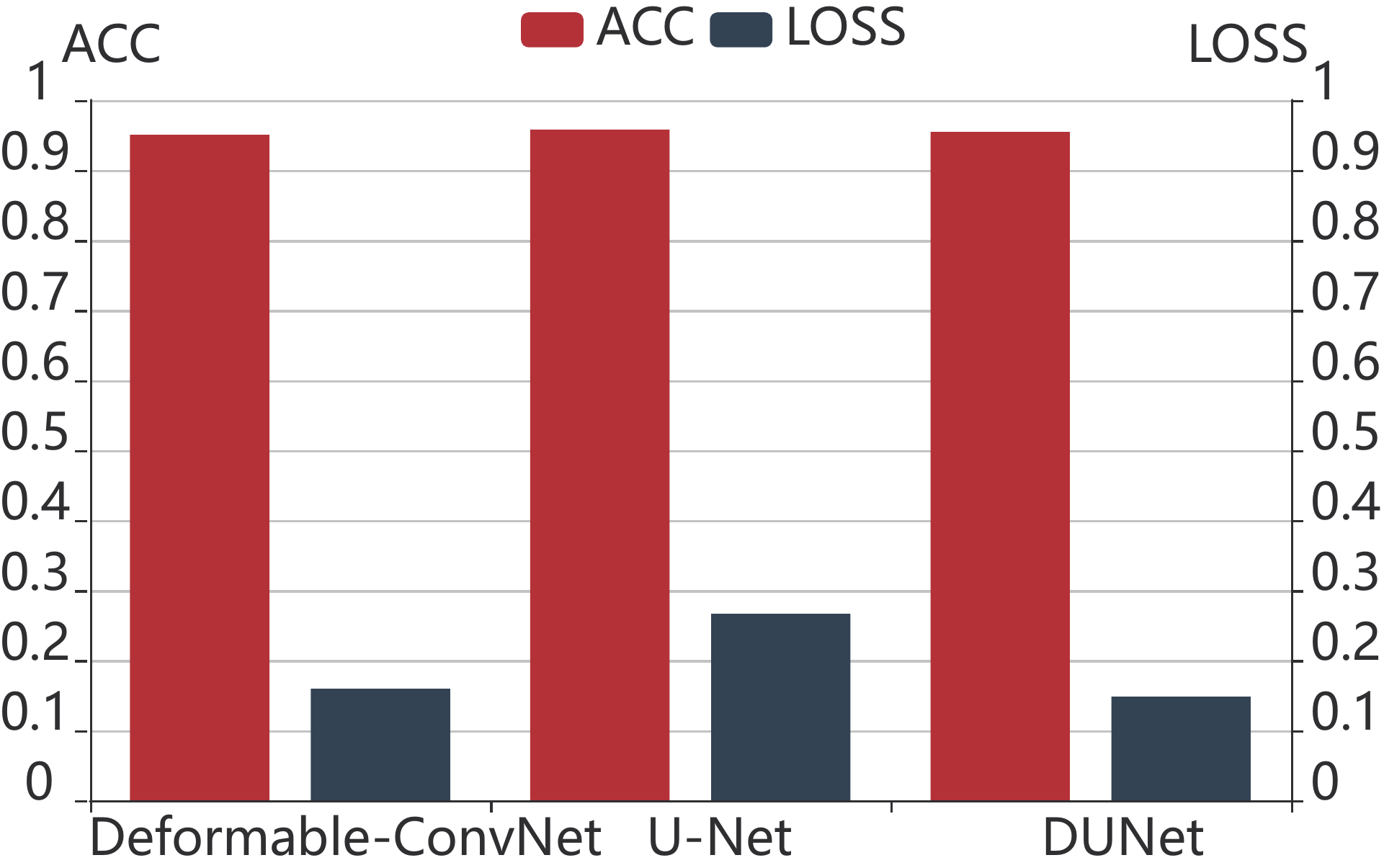}%
\label{stare_performance_svg}}
\hfil
\subfloat[]{\includegraphics[scale=0.3]{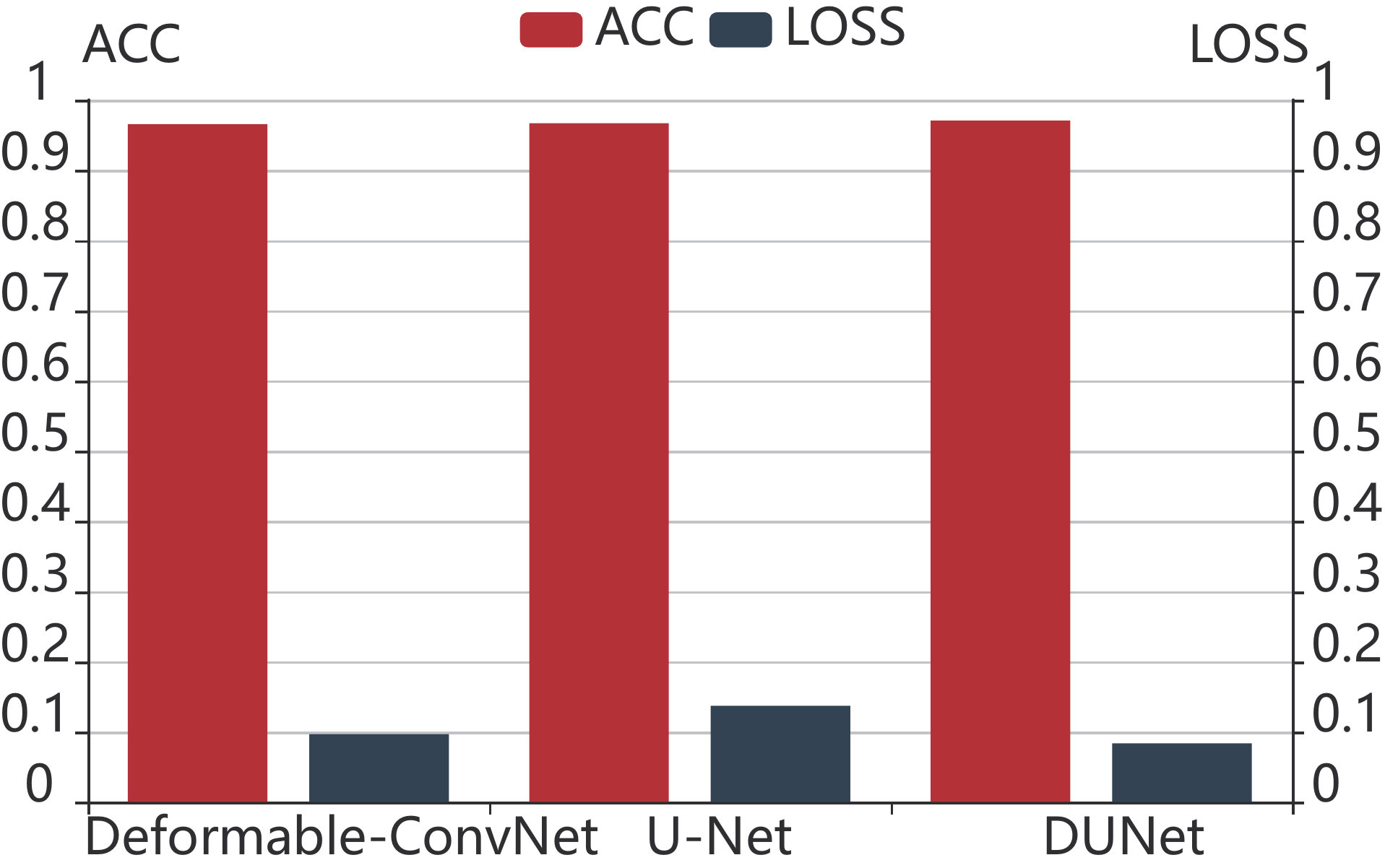}%
\label{chase_performance_svg}}
\caption{Performance comparisons using three models using the validation dataset. (a) validation performance on DRIVE; (b) validation performance on STARE; (c) validation performance on CHASE.}
\label{fig:acc_loss}
\end{figure*}

From Table~\ref{table:performance}, it shows that DUNet achieved the highest validation accuracy of 0.9650 and got the lowest loss value of 0.0919 on DRIVE dataset. On STARE dataset, it had the second highest accuracy and lowest loss value. And on CHASE dataset, it had the highest validation accuracy of 0.9704 and got the lowest loss value of 0.0833. Bar chart of the performance is shown in Fig.~\ref{fig:acc_loss}.

\begin{figure*}
\centering
\subfloat[]{\includegraphics[scale=0.33]{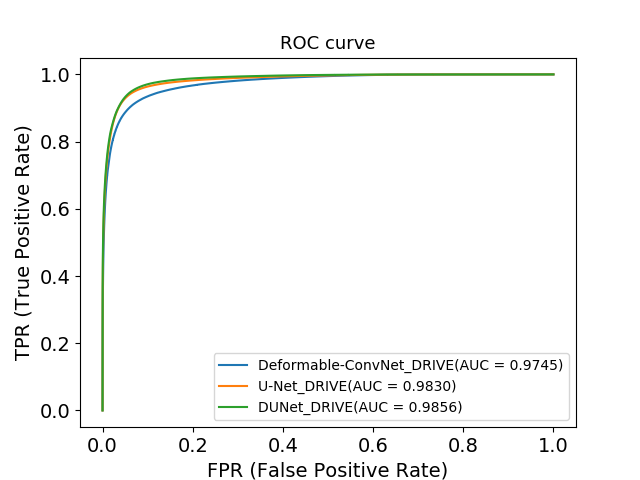}%
\label{drive_roc}}
\hfil
\subfloat[]{\includegraphics[scale=0.33]{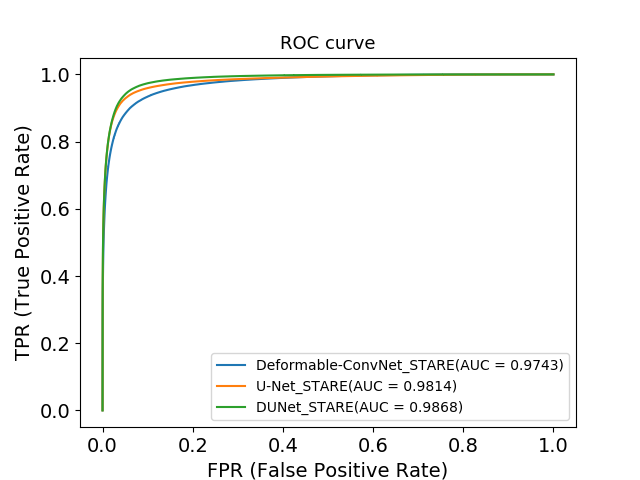}%
\label{stare_roc}}
\hfil
\subfloat[]{\includegraphics[scale=0.33]{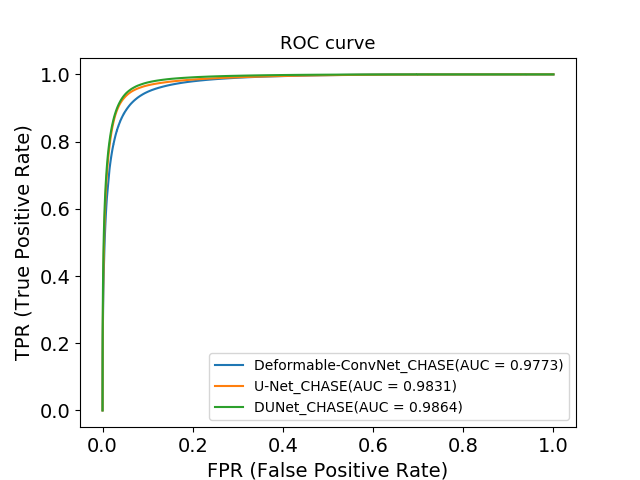}%
\label{chase_roc}}
\caption{ROC curves of different models. (a) ROC curves on DRIVE; (b) ROC curves on STARE; (c) ROC curves on CHASE.}
\label{fig:roc}
\end{figure*}

We further evaluated the model using the test data. PPV, TPR, TNR, ACC, $\mathrm{F_{1}}$-scores, JS and AUC were compared and shown in Table~\ref{table:DRIVE_performance}, Table~\ref{table:STARE_performance} and Table~\ref{table:CHASE_performance}. It shows from the tables that the DUNet achieves the best performance in terms of most of the metrics. To be noticed, the DUNet achieves the highest accuracy among the three models. The global accuracy for Deformable-ConvNet, U-Net, and DUNet is 0.9642/0.9681/0.9697 on DRIVE, 0.9673/0.9705/0.9729 on STARE and 0.9659/0.9728/0.9724 on CHASE, respectively.

\begin{table}[]
\caption{Performance of the three models tested on DRIVE}
\renewcommand\arraystretch{1.3}      
\renewcommand\tabcolsep{1.5pt} 
\begin{center}
\begin{tabular}{cccccccc}
\toprule
\multirow{2}{*}{Models} & \multicolumn{7}{c}{DRIVE}                                                                                                  \\ \cline{2-8} 
                        & PPV             & TPR             & TNR             & ACC             & $\mathrm{F_{1}}$              & JS              & AUC             \\ \hline
Deformable-ConvNet      & 0.8180           & 0.7618          & 0.9837          & 0.9642          & 0.7889          & 0.9642          & 0.9745          \\
U-Net                   & \textbf{0.8795} & 0.7373          & \textbf{0.9903} & 0.9681          & 0.8021          & 0.9681          & 0.9830           \\ 
\textbf{DUNet}                   & 0.8537          & \textbf{0.7894} & 0.9870           & \textbf{0.9697} & \textbf{0.8203} & \textbf{0.9697} & \textbf{0.9856} \\
\bottomrule 
\end{tabular}
\end{center}
\label{table:DRIVE_performance}
\end{table}

\begin{table}[]
\caption{Performance of the three models tested on STARE}
\renewcommand\arraystretch{1.3}      
\renewcommand\tabcolsep{1.5pt} 
\begin{center}
\begin{tabular}{cccccccc}
\toprule
\multirow{2}{*}{Models} & \multicolumn{7}{c}{STARE}                                                                                                  \\ \cline{2-8} 
                        & PPV             & TPR             & TNR             & ACC             & $\mathrm{F_{1}}$              & JS              & AUC             \\ \hline
Deformable-ConvNet      & 0.8447          & 0.7036          & 0.9892          & 0.9673          & 0.7677          & 0.9674          & 0.9742          \\ 
U-Net                   & \textbf{0.9225} & 0.6712          & \textbf{0.9953} & 0.9705          & 0.7770          & 0.9705          & 0.9813          \\
\textbf{DUNet}                   & 0.8856          & \textbf{0.7428} & 0.9920          & \textbf{0.9729} & \textbf{0.8079} & \textbf{0.9729} & \textbf{0.9868} \\
\bottomrule
\end{tabular}
\end{center}
\label{table:STARE_performance}
\end{table}

\begin{table}[]
\caption{Performance of the three models tested on CHASE}
\renewcommand\arraystretch{1.3}      
\renewcommand\tabcolsep{1.5pt} 
\begin{center}
\begin{tabular}{cccccccc}
\toprule
\multirow{2}{*}{Models} & \multicolumn{7}{c}{CHASE}                                                                                                  \\ \cline{2-8} 
                        & PPV             & TPR             & TNR             & ACC             & $\mathrm{F_{1}}$              & JS              & AUC             \\ \hline
Deformable-ConvNet      & 0.7024         & 0.7727          & 0.9786          & 0.9659          &  0.7359          & 0.9659         & 0.9772          \\ 
U-Net                   & \textbf{0.8211} &  0.7124         & \textbf{0.9898} &\textbf{0.9728}          &  0.7629           & \textbf{0.9728}          & 0.9830          \\
\textbf{DUNet}                   & 0.7510         & \textbf{0.8229} & 0.9821          & 0.9724 & \textbf{0.7853} & 0.9724 & \textbf{0.9863} \\
\bottomrule
\end{tabular}
\end{center}
\label{table:CHASE_performance}
\end{table}

\begin{table}[]
\caption{Comparisons against existing approaches on DRIVE dataset}
\renewcommand\arraystretch{1.3}      
\renewcommand\tabcolsep{1.0pt} 
\begin{center}
\begin{tabular}{lccccccc}
\toprule
 Method & Type  & Year   & PPV   & TPR      & TNR   & ACC   & AUC        \\ \hline
 Azzopardi et al.~\cite{azzopardi_trainable_2015}  & STA   & 2015    & -   & 0.7655   & 0.9704 & 0.9442   & 0.9614          \\ 
Li et al.~\cite{Li2015A} & DNN   & 2015    & -   & 0.7569   & 0.9816 & 0.9527   & 0.9738            \\
 Liskowski et al.~\cite{liskowski_segmenting_2016} & DNN   & 2016    & -   & 0.7811   & 0.9807 & 0.9535   & 0.9790            \\ 
 Fu et al.~\cite{fu_deepvessel:_2016}        & DNN        & 2016     & -   & 0.7603   & -      & 0.9523   & -               \\ 
 Dasgupta et al.~\cite{dasgupta_fully_2017}     & DNN     & 2017    & 0.8498 & 0.7691 & 0.9801 & 0.9533   & 0.9744     \\ 
 Roychowdhury et al.~\cite{Roychowdhury2017Blood} & STA & 2017    & -    & 0.7250   & 0.9830 & 0.9520   & 0.9620          \\ 
 Chen et al.~\cite{chen2017a}        & DNN        & 2017  & -    & 0.7426   & 0.9735 & 0.9453   & 0.9516          \\ 
 Alom et al.~\cite{alom_recurrent_2018}        & DNN        & 2018  & -    & 0.7792   & 0.9813 & 0.9556   & 0.9784          \\ 
 \textbf{DUNet}  & \textbf{DNN} & \textbf{2018} & \textbf{0.8537} & \textbf{0.7894} & \textbf{0.9870} & \textbf{0.9697} & \textbf{0.9856} \\ 
\bottomrule
\end{tabular}
\end{center}
\label{table:DRIVE_comparisons}
\end{table}

\begin{table}[]
\caption{Comparisons against existing approaches on STARE dataset}
\renewcommand\arraystretch{1.3}      
\renewcommand\tabcolsep{1.0pt} 
\begin{center}
\begin{tabular}{lccccccc}
\toprule
 Method & Type  & Year   & PPV   & TPR      & TNR   & ACC   & AUC        \\ \hline
 Azzopardi et al.~\cite{azzopardi_trainable_2015}  & STA   & 2015    & -   & 0.7716   & 0.9701 & 0.9497   & 0.9563          \\ 
 Li et al.~\cite{Li2015A} & DNN   & 2015    & -   & 0.7726   & 0.9844 & 0.9628   & 0.9879            \\
 Liskowski et al.~\cite{liskowski_segmenting_2016} & DNN   & 2016    & -   & 0.8554   & 0.9862 & 0.9729   & 0.9928           \\ 
 Fu et al.~\cite{fu_deepvessel:_2016}        & DNN        & 2016     & -   & 0.7412   & -      & 0.9585   & -               \\ 
 Roychowdhury et al.~\cite{Roychowdhury2017Blood} & STA & 2017    & -    & 0.7720   & 0.9730 & 0.9510   & 0.9690          \\ 
 Chen et al.~\cite{chen2017a}        & DNN        & 2017  & -    & 0.7295   & 0.9696 & 0.9449   & 0.9557          \\ 
 Alom et al.~\cite{alom_recurrent_2018}        & DNN        & 2018  & -    & 0.8298   & 0.9862 & 0.9712   & 0.9914          \\ 
 \textbf{DUNet}  & \textbf{DNN} & \textbf{2018} & \textbf{0.8856} & \textbf{0.7428} & \textbf{0.9920} & \textbf{0.9729} & \textbf{0.9868} \\ 
\bottomrule
\end{tabular}
\end{center}
\label{table:STARE_comparisons}
\end{table}

\begin{table}[]
\caption{Comparisons against existing approaches on CHASE dataset}
\renewcommand\arraystretch{1.3}      
\renewcommand\tabcolsep{1.0pt} 
\begin{center}
\begin{tabular}{lccccccc}
\toprule
Method & Type  & Year   & PPV   & TPR      & TNR   & ACC   & AUC        \\ \hline
 Azzopardi et al.~\cite{azzopardi_trainable_2015}  & STA   & 2015    & -   & 0.7585   & 0.9587 & 0.9387   & 0.9487          \\ 
 Li et al.~\cite{Li2015A} & DNN   & 2015    & -   & 0.7507   & 0.9793 & 0.9581   & 0.9716            \\
 Fu et al.~\cite{fu_deepvessel:_2016}        & DNN        & 2016     & -   & 0.7130   & -      & 0.9489   & -               \\ 
 Roychowdhury et al.~\cite{Roychowdhury2017Blood} & STA & 2017    & -    & 0.7201   & 0.9824 & 0.9530   & 0.9532          \\ 
 Alom et al.~\cite{alom_recurrent_2018}        & DNN        & 2018  & -    & 0.7759   & 0.9820 & 0.9634   & 0.9715          \\ 
 \textbf{DUNet}  & \textbf{DNN} & \textbf{2018} & \textbf{0.7510} & \textbf{0.8229} & \textbf{0.9821} & \textbf{0.9724} & \textbf{0.9863} \\ 
\bottomrule
\end{tabular}
\end{center}
\label{table:CHASE_comparisons}
\end{table}

We further evaluated the models using ROC curves, which is shown in Fig.~\ref{fig:roc}. The closer the ROC curve to the top-left border is in the ROC coordinates, the more accurate a model is. It can be seen that the curves of DUNet are the most top-left one among the three models while the Deformable-ConvNet curve is the lowest one of the three. Besides, figures also show that the DUNet obtains the largest area under the ROC curve (AUC).

\begin{figure}
\centering
\includegraphics[scale=0.51]{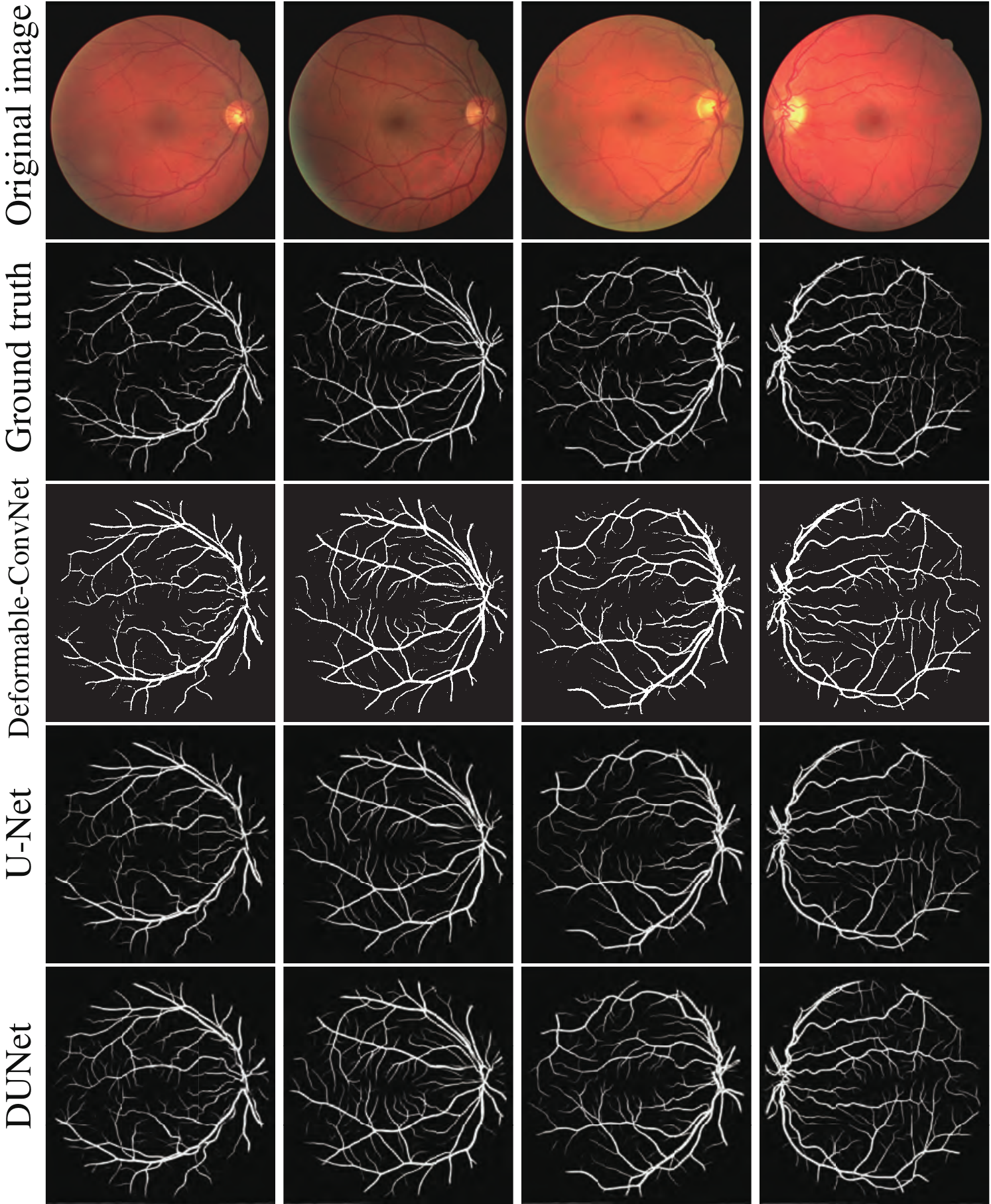}
\caption{Segmentation results using the different models on DRIVE.}
\label{fig:results_drive}
\end{figure}

\begin{figure}
\centering
\includegraphics[scale=0.6]{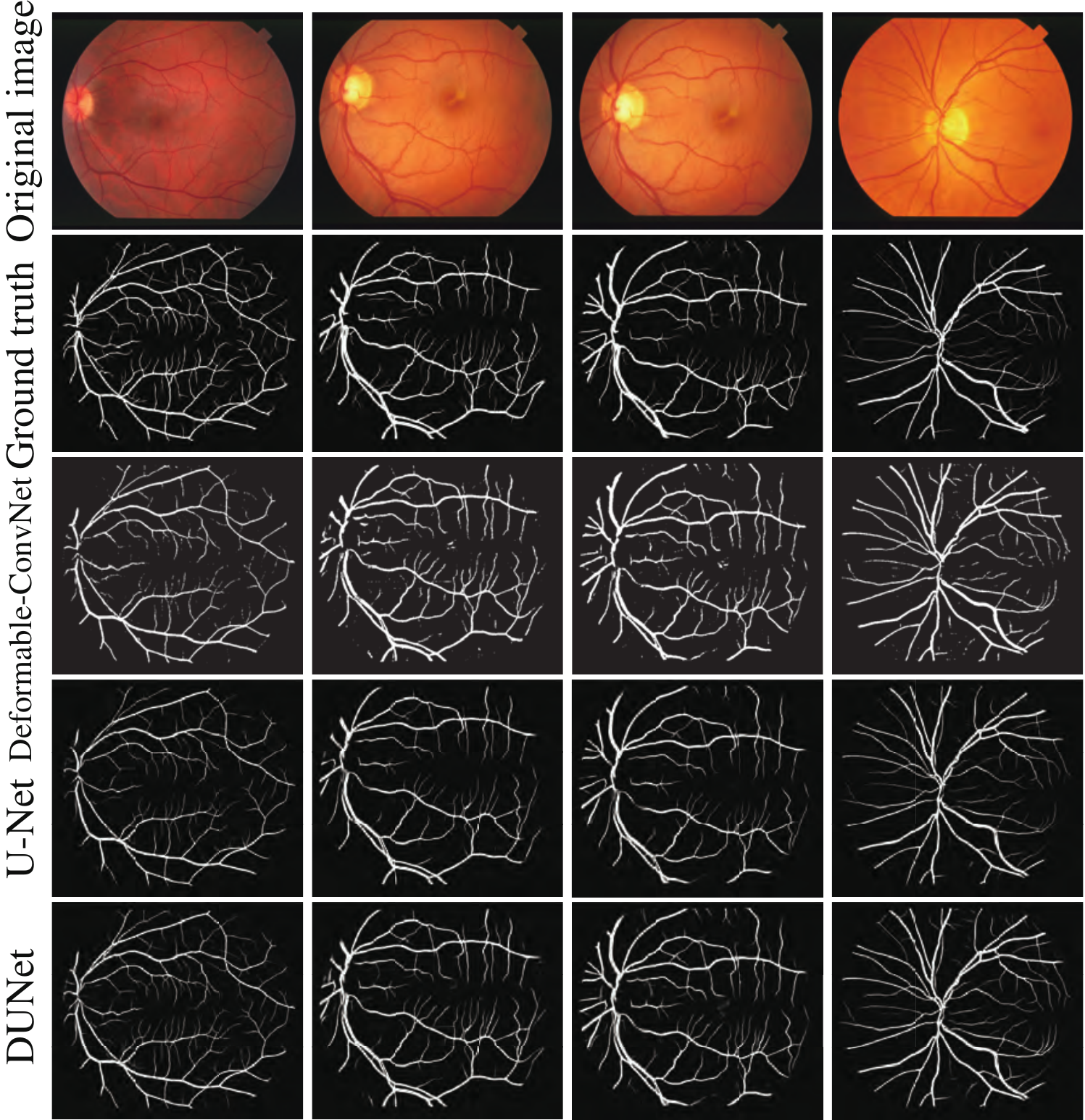}
\caption{Segmentation results using the different models on STARE.}
\label{fig:results_stare}
\end{figure}

\begin{figure}
\centering
\includegraphics[scale=0.58]{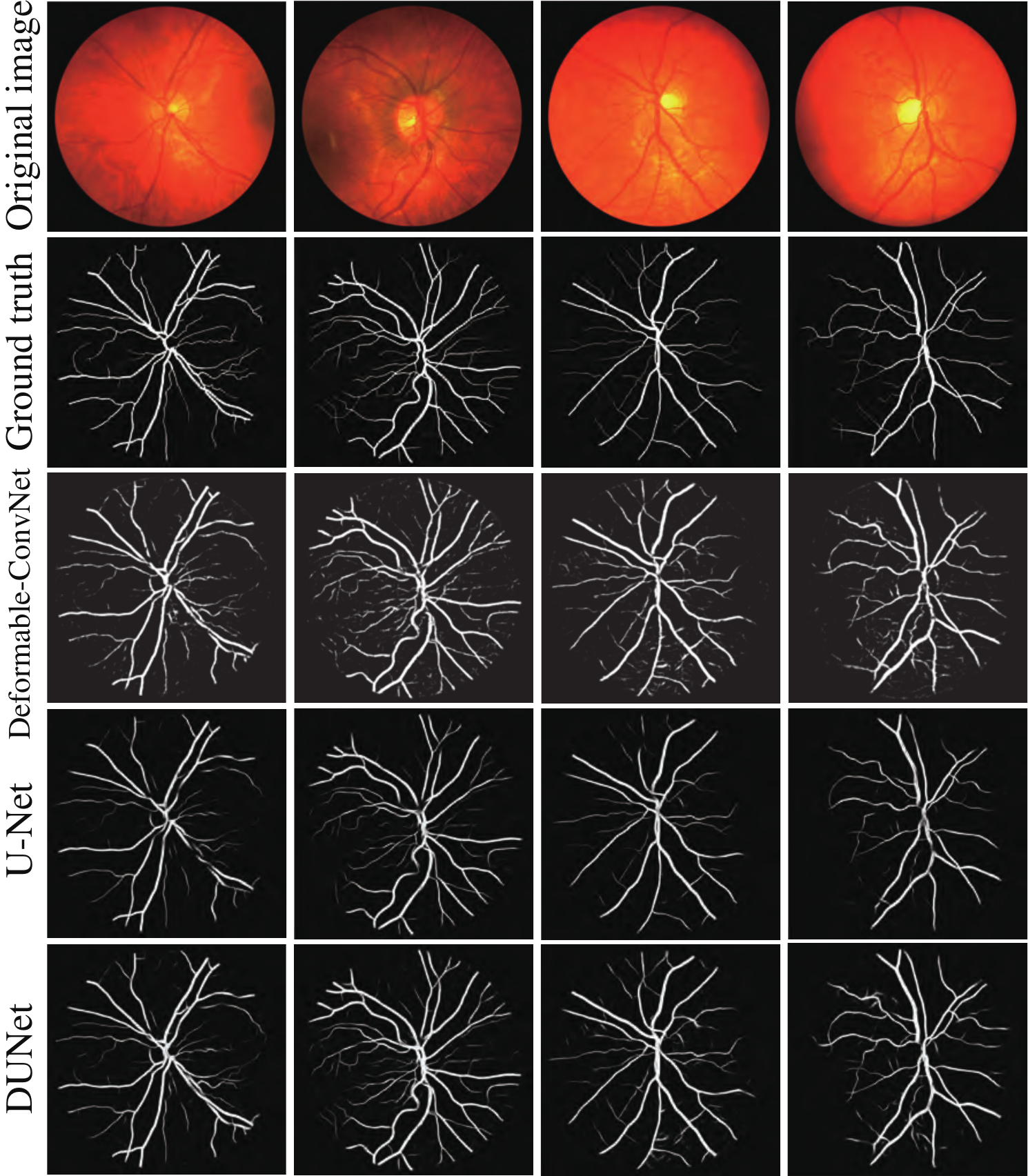}
\caption{Segmentation results using the different models on CHASE.}
\label{fig:results_chase}
\end{figure}

\begin{figure}
\centering
\includegraphics[scale=0.35]{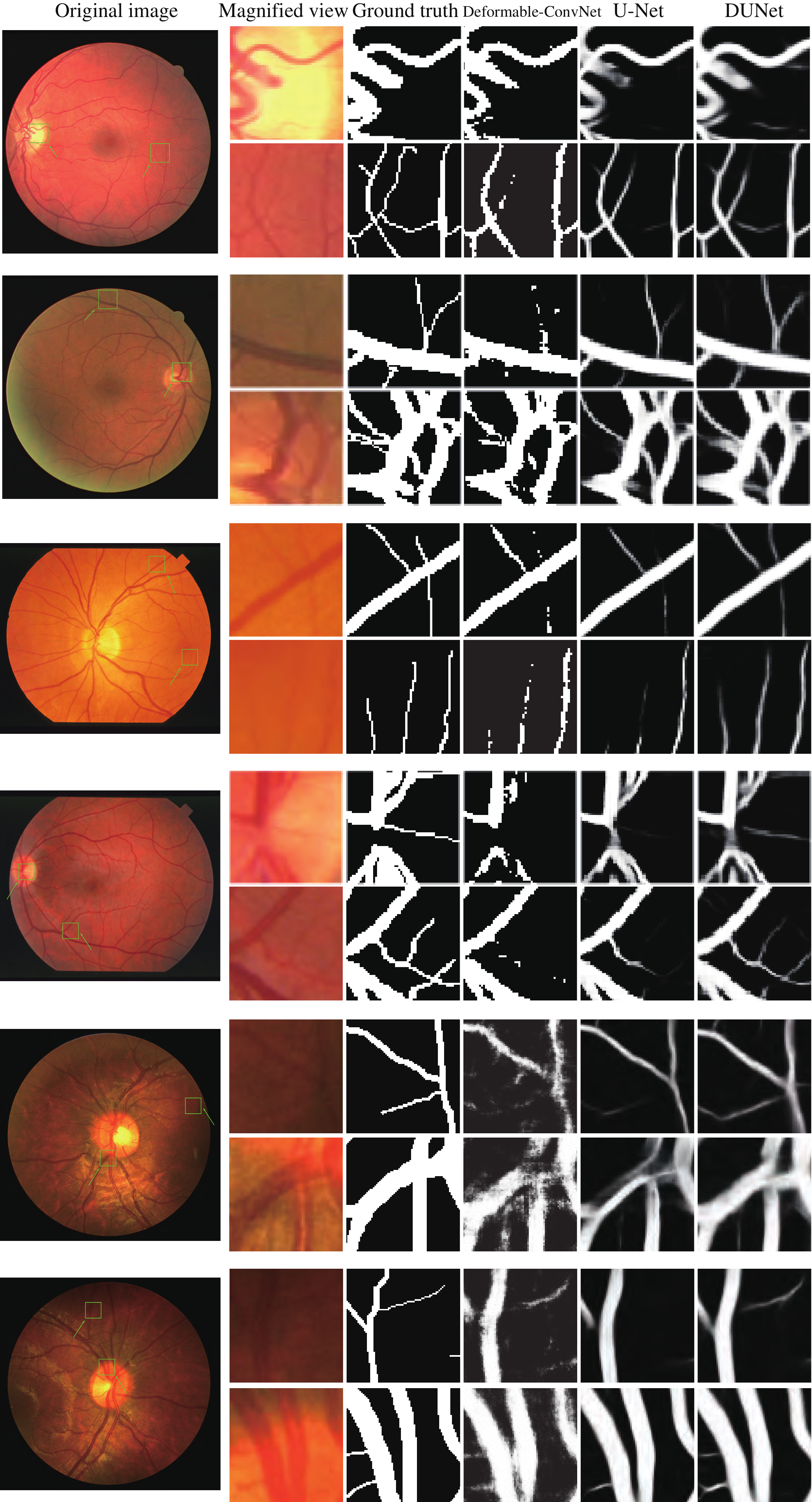}
\caption{Magnified view of green-boxed patches predicted by different models on DRIVE (two rows above), STARE (two rows middle) and CHASE (two rows below).
}
\label{fig:result_magnified}
\end{figure}

\begin{figure}
\centering
\includegraphics[scale=0.44]{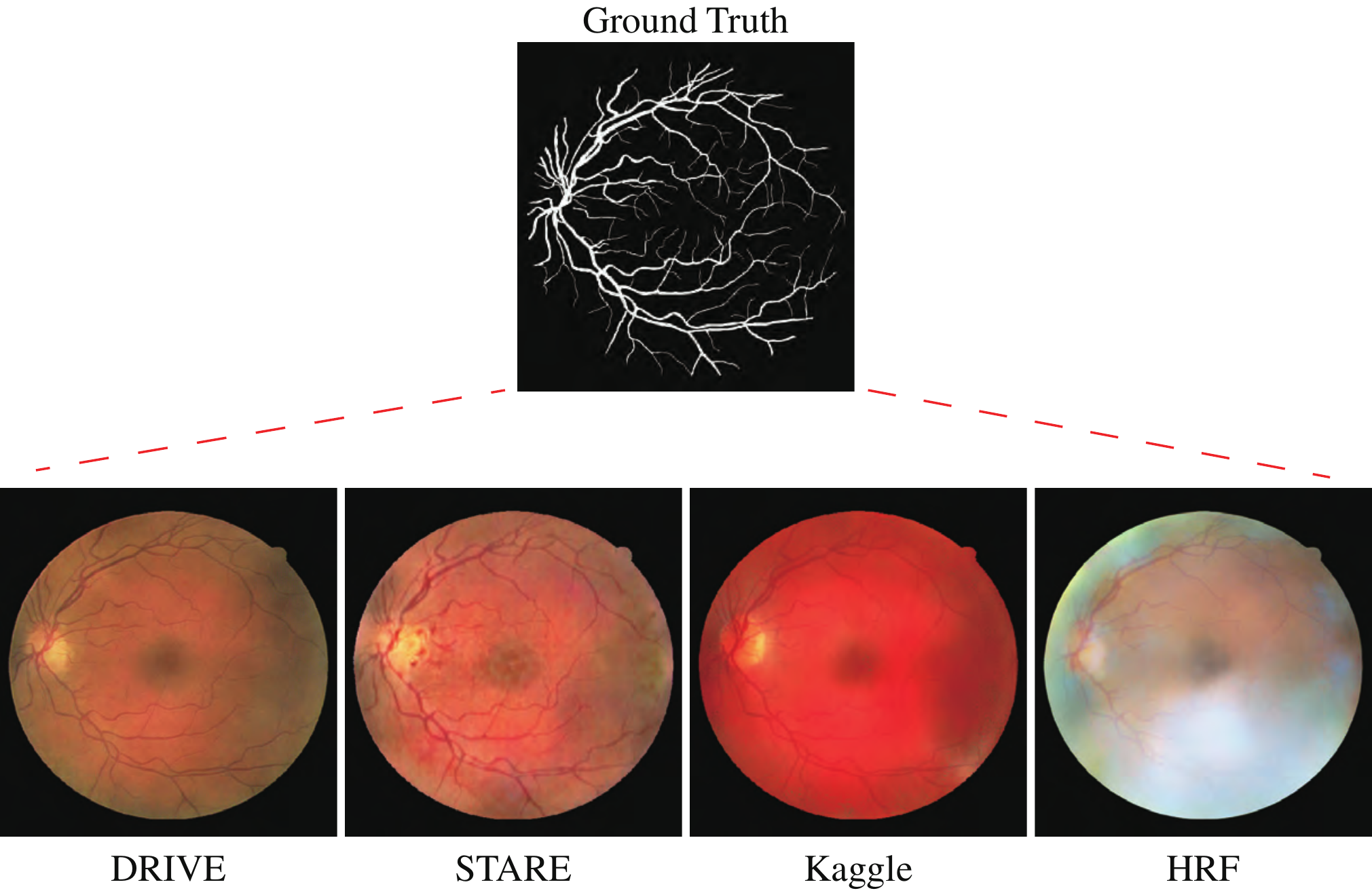}
\caption{Four distinct style of retinal images synthesized by generative adversarial nets.
}
\label{fig:synthe_intro}
\end{figure}

\begin{figure*}
\centering
\includegraphics[scale=0.5]{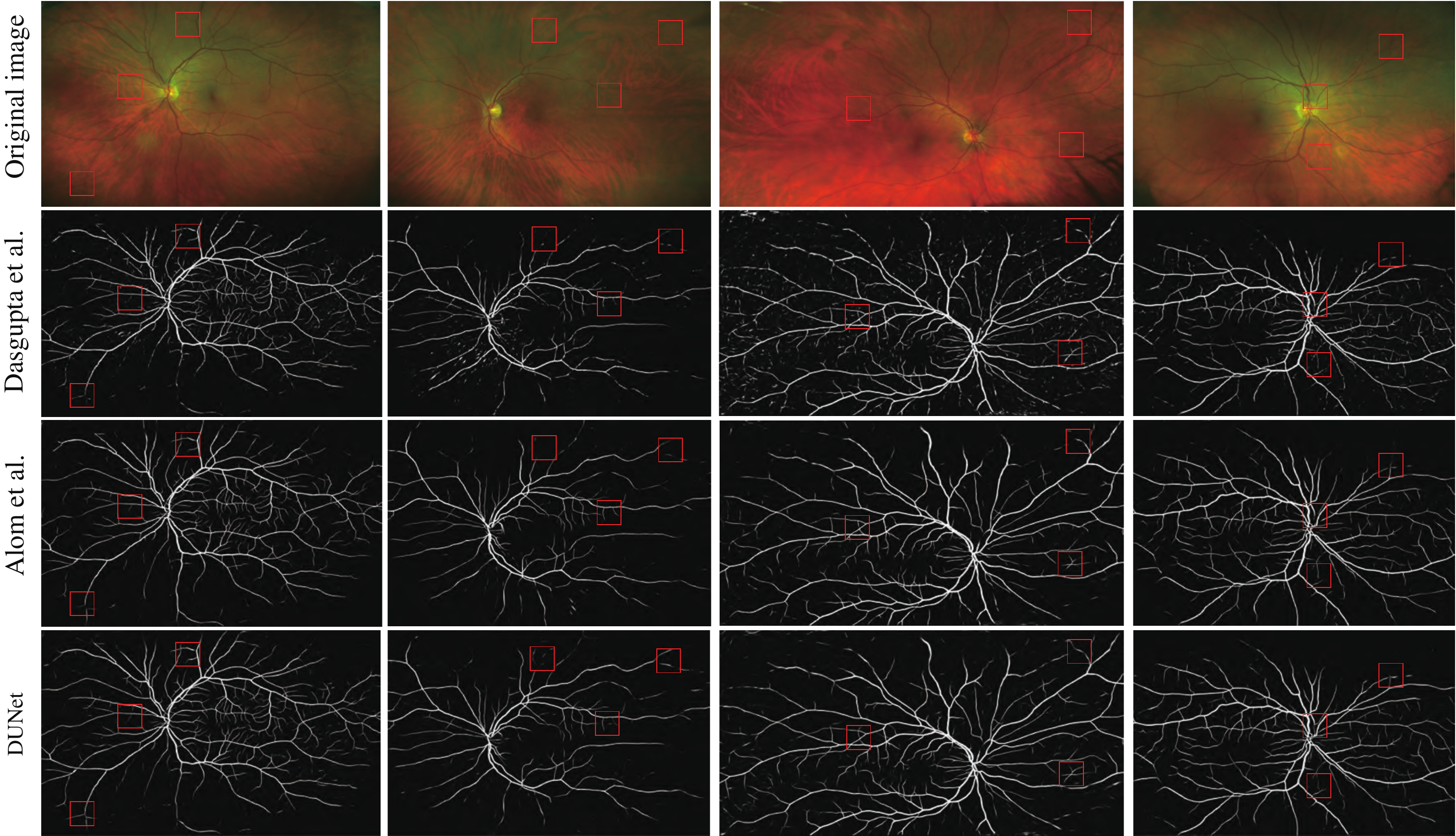}
\caption{Detailed view of four images on WIDE. Red boxes show segmentation cases that DUNet perform better than the other two methods.
}
\label{fig:compare_with_other_on_wide}
\end{figure*}

\subsection{Retinal vessel segmentation results}
We display the retinal vessel segmentation results in Fig.~\ref{fig:results_drive}, Fig.~\ref{fig:results_stare} and Fig.~\ref{fig:results_chase}. From figures, it can be observed that DUNet produces more distinct vessel segmentation results. The proposed DUNet can detect weak vessel or vessels that are tied up which may be lost in U-Net and Deformable-ConvNet, thus it is more powerful to preserve more details.

We show the details of the segmentation results of the three models in Fig.~\ref{fig:result_magnified}, it shows the local magnification view of vascular junction, where several vessels are tied up and close to each other, and tiny vessels of DRIVE, STARE and CHASE respectively. Due to the complicated vascular tree, segmentation algorithms are difficult to proceed precisely with such complicated structures. In the junction region of vessel, Deformable-ConvNet and U-Net extracted coarse information due to the limitation of network. It is worth mentioning that Deformable-ConvNet extracted more detailed vessel than U-Net in some junction regions, which showed its ability to capture retinal vessels of various shapes. With the help of deformable convolutional blocks, the DUNet successfully segmented the tied vessels. In the tiny vessel regions, U-Net showed its limitation in handling details. However, Deformable-ConvNet picked them up somewhere. As a result, the DUNet got a desiring segmentation results in those tiny and weak vessels.

With this structure, the DUNet is able to distinguish different vessels and present a better performance than the other models. Experimental results arrival at a conclusion that DUNet architecture has a more desirable performance in dealing with complicated and weak vessel structures among the three models.

\subsection{Comparison against existing methods}
We also compared our method with several state-of-the-art approaches. Among them, some are standard segmentation algorithms (denoted with STA) while the others are all based on deep neural networks (denoted with DNN). Table~\ref{table:DRIVE_comparisons},~\ref{table:STARE_comparisons}, \ref{table:CHASE_comparisons} summarize the type of algorithm, year of publication, and performance on DRIVE, STARE and CHASE dataset. From the results, it shows that DUNet architecture performs the best among those methods on DRIVE and CHASE. It achieves the highest global accuracy of 0.9697/0.9724 and the highest AUC of 0.9856/0.9863 with a small quantity of training samples, which shows that the DUNet exhibits state-of-the-art performance comparing both standard segmentation methods and deep neural network based methods. Although DUNet performs not better than Liskowski et al.'s method~\cite{liskowski_segmenting_2016} and Alom et al.'s method~\cite{alom_recurrent_2018} on STARE, DUNet uses less training patch samples than their methods while reaches a desiring results.

Additionally, we compared our method with  Dasgupta et al.'s method and Alom et al.'s method on the other two datasets for qualitative and quantitative analysis. The first dataset named WIDE, used for tree topology estimation, contains 15 high-resolution, wide-field, RGB images. Each retinal image was taken from a different individual and captured as an un-compressed TIFF file at the widest setting~\cite{Estrada2015Tree}. The WIDE dataset does not contain ground truth for retinal vessel segmentation. We used the WIDE for qualitative analysis and compared the proposed method with the other two methods. The second dataset (denoted with SYNTHE) is synthesized from generative adversarial nets~\cite{zhao2018synthesizing}. The dataset contains 20 retinal images at $565 \times 584$ resolution, which includes DRIVE~\cite{staal_ridge-based_2004}, STARE~\cite{hoover_locating_1998}, Kaggle and HRF~\cite{Kohler2013Automatic} style. Each of these styles contains 5 retinal images generated by 5 corresponding ground truth images. Fig.~\ref{fig:synthe_intro} shows the SYNTHE dataset, four distinct style retinal images are generated from the same vessel map.

To show the generalization of these three models, we used the weights well-trained on DRIVE and predicted on WIDE and SYNTHE dataset. We preprocessed and cropped these images in patches in the same way. From Fig.~\ref{fig:compare_with_other_on_wide}, it qualitatively indicates that DUNet produces competitive results. 

To further validate quantitatively the performance of these models, we also used the well-trained weights from DRIVE and tested on the SYNTHE datasets. We mixed the four distinct style images together, preprocessed and cropped SYNTHE in patches in the same way. The performances of three models are summarized in Table~\ref{table:SYNTHE_performance}, which prove quantitatively that the DUNet gets the best performance among all these three models overall. 

\begin{table}[]
\caption{Performances of the three models tested on SYNTHE using well-trained weights on DRIVE}
\renewcommand\arraystretch{1.3}      
\renewcommand\tabcolsep{1.5pt} 
\begin{center}
\begin{tabular}{lccccccc}
\toprule
\multirow{2}{*}{Models} & \multicolumn{7}{c}{SYNTHE}                                                                                                  \\ \cline{2-8} 
                        & PPV             & TPR             & TNR             & ACC             & $\mathrm{F_{1}}$              & JS              & AUC             \\ \hline
 Dasgupta et al.~\cite{dasgupta_fully_2017}       & 0.8485          & 0.7660         & 0.9868          & 0.9675          & 0.8052          & 0.9675          & 0.9822          \\ 
Alom et al.~\cite{alom_recurrent_2018}                   & 0.8509 & 0.7728          & 0.9870 & 0.9682          & 0.8100           & 0.9682          & 0.9831          \\
DUNet                   & \textbf{0.8537}          & \textbf{0.7894} & \textbf{0.9870}          & \textbf{0.9697} & \textbf{0.8203} & \textbf{0.9697} & \textbf{0.9855} \\
\bottomrule
\end{tabular}
\end{center}
\label{table:SYNTHE_performance}
\end{table}

\section{Conclusion}
\label{sec:conclusion}
Deep neural networks, which uses hierarchical layers of learned features to accomplish high-level tasks, has been applied to a wide range of medical processing tasks. In this study, we propose a fully convolutional neural network, named DUNet to handle the retinal vessel segmentation task in a pixel-wise manner. DUNet is an extension of the U-Net with convolutional layers replaced by the deformable convolution blocks. With the symmetric U-shape architecture, DUNet is designed to capture context by the encoder and enable precise localization by the decoder through combining the low-level feature maps with the high-level ones. It also allows the context being propagated to the higher resolution layers through a larger number of feature channels in the upsampling part. Furthermore, with the deformable convolution blocks, DUNet is able to capture the retinal blood vessels at various shapes and scales by adaptively adjusting the receptive fields according to the vessels' scales and shapes. By adding offsets to the regular sampling grids of standard convolution, the receptive fields are deformable and augmented. While it does bring some extra costs of computation resources from convolution offset layer. In order to test the performance of the proposed network, we have trained Deformable-ConvNet and U-Net from scratch for comparison. This is also the first time that DUNet being used to conduct the retinal segmentation. Besides, a comparison with several standard segmentation algorithms and some other deep neural network based approaches are introduced here. We train and test the models on three public datasets: DRIVE, STARE and CHASE\_DB1. To validate the generalization of our model, we tested the DUNet on WIDE and SYNTHE datasets, and analyze qualitatively and quantitatively. Results show that with the help of deformable convolutional blocks, more detailed vessels are extracted, and the DUNet exhibits state-of-the-art performance in segmenting the retinal vessels.

In the future, more retinal vessel data will be incorporated to validate the proposed end-to-end model. We also plan to extend our DUNet architecture to three dimensions, aiming to obtain more accurate results in medical image analysis tasks.

\section*{Acknowledgment}

This work is supported by the Science and Technology Program of Tianjin, China [Grant No. 16ZXHLGX00170], the National Key Technology R\&D Program of China [Grant No. 2015BAH52F00] and the National Natural Science Foundation of China [Grant No. 61702361].

\ifCLASSOPTIONcaptionsoff
  \newpage
\fi

\bibliographystyle{IEEEtran}
\bibliography{bare_jrnl}

\end{document}